\documentclass[10pt,twocolumn,letterpaper]{article}

\usepackage[pagenumbers]{cvpr} 

\usepackage[dvipsnames]{xcolor}
\usepackage{color,colortbl}
\usepackage{todonotes}
\usepackage{pict2e,multirow}
\usepackage[pagebackref,breaklinks,colorlinks,citecolor=cvprblue]{hyperref}
\usepackage{orcidlink}
\usepackage{bm}
\usepackage{microtype}
\usepackage{overpic}
\definecolor{cvprblue}{rgb}{0.21,0.49,0.74}

\newcommand{\taskname}{VoF3DIS\xspace}
\newcommand{\ovname}{OV3DIS\xspace}
\newcommand{\ourmethod}{PoVo\xspace}

\newcommand{\vl}{vision-language\xspace}
\newcommand{\llava}{LLaVA\xspace}
\newcommand{\prompt}{``\texttt{List the object names in the scene}''\xspace}

\definecolor{OpenVoc}{RGB}{229,255,214}
\definecolor{VocFree}{RGB}{214,214,255}
\definecolor{SemUnd}{RGB}{213,232,212}
\definecolor{SupMet}{RGB}{218,232,252}

\DeclareMathOperator*{\argmax}{argmax} 

\definecolor{OpenVoc}{RGB}{229,255,214}
\definecolor{VocFree}{RGB}{214,214,255}
\definecolor{SemUnd}{RGB}{218,232,252}
\definecolor{GeoUnd}{RGB}{213,232,212}

\def\paperID{26} 
\def\confName{3DV\xspace}
\def\confYear{2025\xspace}

\title{Vocabulary-Free 3D Instance Segmentation with Vision-Language Assistant}

\author{Guofeng Mei \hspace{3mm} Luigi Riz \hspace{3mm} Yiming Wang \hspace{3mm} Fabio Poiesi\\
 Fondazione Bruno Kessler \\
Via Sommarive, 18, 38123 Trento, Italy \\
{\tt\small \{gmei,luriz,ywang,poiesi\}@fbk.eu}
}

\begin{document}
\twocolumn[{%
\renewcommand\twocolumn[1][]{#1}%
\maketitle
\vspace{-2mm}
\begin{center}
    \vspace{-4mm}
    \centering
    \captionsetup{type=figure}
    \includegraphics[width=\linewidth]{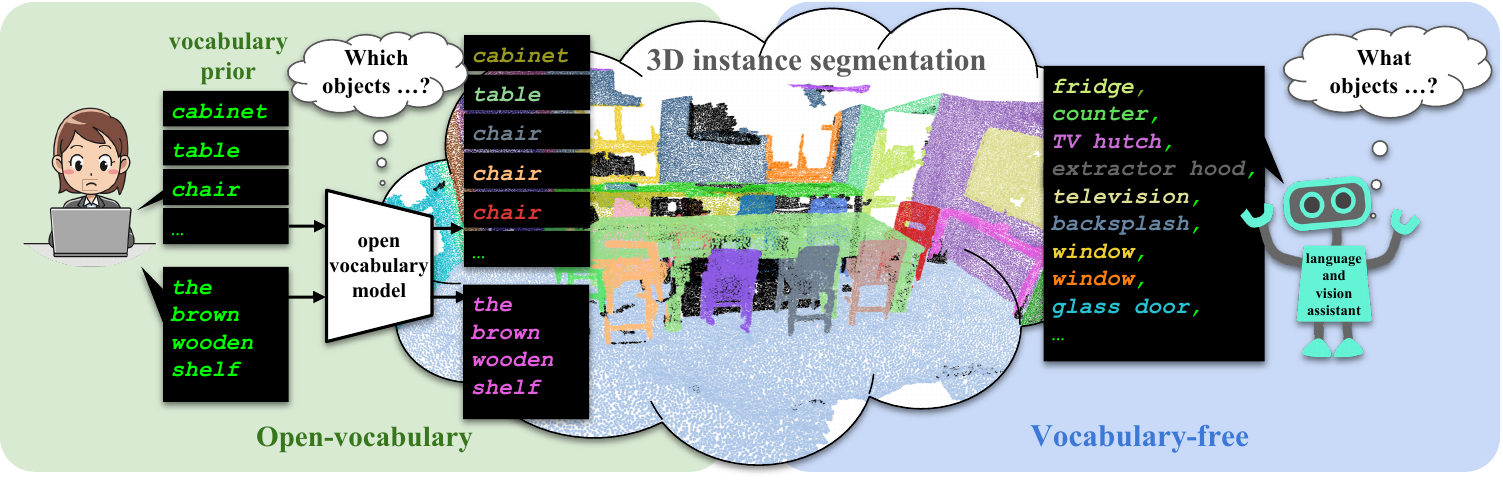}
    \vspace{-7mm}
    \captionof{figure}{We introduce a vocabulary-free approach to address 3D instance segmentation that leverages \vl assistants, moving beyond the limitations of open-vocabulary approaches.
Left: `Open-vocabulary', where 3D instances are segmented using the user-specified restricted lexical scope, \ie,~the `vocabulary prior'. 
Right: `Vocabulary-free', our approach can understand scenes without relying on vocabulary prior, and autonomously recognizes a wide-range of objects, \eg~backsplash and TV hutch.}
\label{fig:teaser}
\end{center}%
}]
\begin{abstract}
\vspace{-5mm}
Most recent 3D instance segmentation methods are open vocabulary, offering a greater flexibility than closed-vocabulary methods. Yet, they are limited to reasoning within a specific set of concepts, \ie, the vocabulary, prompted by the user at test time. 
In essence, these models cannot reason in an open-ended fashion, \ie, answering ``\texttt{List the objects in the scene.}''
We introduce the first method to address 3D instance segmentation in a setting that is void of any vocabulary prior, namely a vocabulary-free setting.
We leverage a large \vl assistant and an open-vocabulary 2D instance segmenter to discover and ground semantic categories on the posed images. 
To form 3D instance masks, we first partition the input point cloud into dense superpoints, which are then merged into 3D instance masks. We propose a novel superpoint merging strategy via spectral clustering, accounting for both mask coherence and semantic coherence that are estimated from the 2D object instance masks.
We evaluate our method using ScanNet200 and Replica, outperforming existing methods in both vocabulary-free and open-vocabulary settings. \footnote{Project page: \url{https://gfmei.github.io/PoVo}}
\end{abstract}    

\vspace{-2mm}
\section{Introduction}\label{sec:intro}
\vspace{-2mm}
3D instance segmentation (3DIS) is a challenging research problem that demands instance-level semantic understanding and precise object delineation.
Given a 3D scene, 3DIS aims to produce a set of binary masks associated with their semantic labels, where each of them correspond to an object instance.
Traditional methods addressing 3DIS follow a closed-vocabulary paradigm~\cite{chen2021hierarchical, zhong2022maskgroup,vu2022softgroup,schult2023mask3d}, where the set of semantic categories that can be encountered at test time is the same as that seen at training time.
With advances in \vl models, the 3DIS literature has rapidly evolved from closed-vocabulary methods to \textit{open-vocabulary} methods \cite{takmaz2023openmask3d,nguyen2024open3dis}, where the semantic categories at test time can be different from those seen during training.
Open-vocabulary 3DIS (\ovname) methods focus on obtaining i) instance-level 3D masks by using class-agnostic 3D segmentation methods (\eg Mask3D~\cite{schult2023mask3d} or superpoints~\cite{felzenszwalb2004efficient}), and ii) the corresponding text-aligned mask representation by aggregating text-aligned visual representations from posed images, \eg obtained by CLIP. 
Scene semantics are obtained by assuming certain \textit{vocabulary prior} (left-hand side of \cref{fig:teaser}): either in the form of a large finite set of categories~\cite{peng2023openscene, tai2024open}, akin to answer ``\texttt{Choose among the \textit{vocabulary}, which object do these points correspond to?}'', or in the form of human query with specified object instance~\cite{takmaz2023openmask3d}, akin to answer ``\texttt{Do these points correspond to \textit{an armchair with floral print}?}''.

\textit{What if the vocabulary prior is unknown at the inference time? }
Such scenarios can occur in assistive robotic applications, especially when the scene semantics are dynamically evolving, \eg inclusion or replacement of instances that are not defined in the vocabulary or unknown to the user, such as an uncommon utensil or a specific painting.
Thus, it would be ideal to empower 3DIS methods with the capability in answering open-ended questions, such as ``\texttt{What object do these points correspond to?}''. 
Inspired by the naming convention in the image recognition field~\cite{conti2023vocabularyfree}, we term such setting as \textbf{Vo}cabulary-\textbf{F}ree \textbf{3D} \textbf{I}nstance \textbf{S}egmentation (\textbf{\taskname}), which further extends the \ovname by lifting the need of any vocabulary prior.
Formally, \taskname aims to segment all object instances in a point cloud, with their semantic labels associated, without relying on any vocabulary prior, \eg~pre-defined vocabulary or user-specified query at test time (right-hand side of \cref{fig:teaser}).

Previous works in such a vocabulary-free setting primarily focus on image recognition and tackle the challenge of obtaining relevant categories in a vast semantic space. 
Different methods have been explored to form candidate vocabularies, including retrieving from a multimodal web-scale database~\cite{conti2023vocabularyfree} or querying from \vl assistants~\cite{liu2024democratizing}, \eg \llava~\cite{liu2023llava}.
\taskname instead focuses on understanding the 3D scene at the instance level, which presents significant new challenges, as the scale of 3D data is not comparable to that of 2D scene understanding.
Moreover, it not only faces the hurdle of a vast semantic space but also demands point-level semantic representation - \taskname technically nontrivial aspect, as existing large \vl models typically process the input image as a whole~\cite{liu2023llava, blip2}.

We propose the first point cloud vocabulary-free instance segmentation method, \ourmethod, which can semantically understand 3D instance masks utilizing a \vl assistant~\cite{Liu_2024_CVPR}. 
\ourmethod is zero-shot and does not require any training on either 2D or 3D data. Instead, it leverages a vision-language assistant and a visual grounding model to obtain localized scene semantics from 2D posed images. 
Such semantic representations are then lifted to 3D to merge and refine the 3D masks that are initially segmented based on geometric features. 
Specifically, for each posed image, \ourmethod prompts the \vl assistant to retrieve the list of objects present in the scene from the answer, forming a \textit{scene vocabulary}. 
For the 3D instance proposal, \ourmethod first segments the 3D scene into superpoints via graph cut based on geometric features. Then we leverage a visual grounding model, \eg anchored SAM \cite{ren2024grounded}, to obtain semantically aware object masks in posed images using the scene vocabulary.
Those masks and their semantic labels are then used to guide the merging process of superpoints towards 3D instance masks via spectral clustering, using a superpoint affinity matrix based on both spatial and semantic information. 
Lastly, \ourmethod tags such 3D proposals using both vision and text information.
We evaluated \ourmethod on two 3D scene datasets: ScanNet200~\cite{scannet200} and Replica~\cite{straub2019replica}. 
To assess this newly introduced setting (\taskname), we also BERT score~\cite{Zhang2020} with average precision metrics to address the challenge of labeling points with concepts from a vast semantic space. 
Our results demonstrate that \ourmethod outperforms recent approaches adapted to the \taskname setting. 
Moreover, \ourmethod also outperforms competitors in the open-vocabulary setting, validating its robust design in managing a large set of semantic concepts. In summary, our contributions include:
\begin{itemize}[noitemsep,nolistsep,leftmargin=*]
\item introducing the new vocabulary-free task for 3D instance segmentation;
\item proposing \ourmethod, the first method to transfer visually-grounded concepts identified by large vision and language models to point clouds for 3D instance segmentation;
\item proposing a novel 3D mask formation strategy that merges over-segmented superpoints into instance masks using spectral clustering, considering both semantic consistency and mask coherence.
\end{itemize}

\section{Related work}

\noindent\textbf{Vocabulary-free models.}
Conti et al.~\cite{conti2023vocabularyfree} pioneered the vocabulary-free setting, that is assigning ``\textit{an image to a class that belongs to an unconstrained language-induced semantic space at test time, without a vocabulary}''. 
Their method, named CaSED, retrieves captions from a database that are semantically closer to the input image. 
From these captions, candidate categories are extracted through text parsing and filtering. 
CaSED estimates the similarity score between the input image and each candidate category using CLIP, leveraging both visual and textual information, to predict the best matching candidate. 
Subsequent works remain mostly in the image domain, \eg by extending vocabulary-free image classification to semantic image segmentation~\cite{conti2024}, and exploring \vl assistants for fine-grained image classification~\cite{liu2024democratizing}.
To the best of our knowledge, \taskname is the first that extends such vocabulary-free setting to 3D instance segmentation.
Instead of retrieving from a web-scale database, \ourmethod leverages a \vl assistant to obtain relevant semantic concepts for the target 3D scene, making it more flexible and versatile.

\noindent\textbf{Open-vocabulary 3D scene understanding.}
Recent advancements in 3D scene understanding mostly involve the adaptation of Vision-Language Models (VLMs) to the 3D domain, enabling semantic understanding in the open-vocabulary setting, \ie, being able to recognize objects within a wide-range vocabulary, no longer being constrained by the closed-set at training time. 
A common pathway is to transfer visual representation features of multi-view images to 3D points via pixel-to-point correspondences, which can then be used to either train text-aligned 3D encoders~\cite{peng2023openscene,jiang2024open}, or to address open-vocabulary 3D scene understanding tasks directly~\cite{jatavallabhula2023conceptfusion}. In the following, we discuss recent open-vocabulary methods designed for 3D scene semantic segmentation and instance segmentation.

\textit{3D semantic segmentation} methods aim to obtain point-level representation that are aligned with text. 
PLA~\cite{ding2023pla} leverages image captioning models to form hierarchical 3D-caption pairs and employs contrastive learning to align point-level features with textual representation.
OpenScene~\cite{peng2023openscene} learns a 3D network through distillation and uses open-vocabulary 2D segmentation approaches, such as OpenSeg \cite{ghiasi2022openseg} and LSeg \cite{li2022lseg}, to extract pixel-level features from posed images. 
Differently, OV3D~\cite{jiang2024open} prompts \vl assistant to generate semantic classes with detailed descriptions from multi-view images. 
Text with enriched semantics are then anchored to pixels via image segments obtained with class-agnostic segmentation method, \eg, SAM~\cite{kirillov2023sam}. OV3D~\cite{jiang2024open} further trains a 3D encoder to align with the pre-trained text encoder for open-vocabulary semantic segmentation.
ConceptFusion~\cite{jatavallabhula2023conceptfusion} transfers visual features to 3D points in a training-free fashion. It employs a class-agnostic image segmentation method, like SAM \cite{kirillov2023sam}, to localize regions containing objects in images and utilizes a vision encoder, such as CLIP \cite{Radford2021clip}, to generate pixel-level features, which can be lifted to 3D points for open-vocabulary scene understanding.

\textit{3D instance segmentation} methods aim to obtain instance-level 3D masks with associated text-aligned per-mask representation. OpenMask3D~\cite{takmaz2023openmask3d} is a prior method that features open-vocabulary 3D instance segmentation. OpenMask3D relies on Mask3D~\cite{schult2023mask3d}, a class-agnostic 3D instance mask generator to generate 3D instance masks. Each 3D mask is projected to multi-view images to locate corresponding visual regions and obtain text-aligned representation with CLIP encoder. Finally, it matches per-mask representations with textual representation of user-specified queries for segmenting instances.
OVIR-3D~\cite{lu2023ovir} processes multi-view images with an off-the-shelf open-vocabulary 2D detector, Detic~\cite{zhou2022detecting}, to produce 2D region proposals associated with text-aligned features, which are then aggregated to 3D points, forming 3D instance representation for query.
Instead of using a class-agnostic 3D segmenter to generate instance masks~\cite{takmaz2023openmask3d}, recent works~\cite{yin2024sai3d,tai2024open,zhang2024recognize} partition the 3D scene into superpoints and progressively merge them into 3D instance masks with 2D guidance. Finally, they assign a semantic label to each mask or refer to any mask using user-specified text based on CLIP.
For example, SAI3D~\cite{yin2024sai3d} builds a sparse affinity matrix that captures pairwise similarity based on the 2D masks generated by SAM to merge superpoints. Open-vocabulary 3DIS is then achieved by finding the most overlapping 3D mask with the area that are projected from 2D masks obtained by OVSeg~\cite{liang2023open} given the text query.
OVSAM3D~\cite{tai2024open} project the superpoints onto 2D posed images to serve as point prompts to guide SAM for image segmentation. The image segments are then projected back to 3D for refining the 3D masks. Semantic labels are obtained from an open-vocabulary tagging method, RAM~\cite{zhang2024recognize} with a vocabulary of about 6,400 categories. OVSAM3D also leverages a language assistant (ChatGPT) to filter out scene-irrelevant concepts. Semantic categories can then be anchored with CLIP by comparing the visual embeddings of SAM segmented 2D crops against the text embeddings of RAM-obtained tags.
Open3DIS~\cite{nguyen2024open3dis} further combines the class-agnostic 3D instance segmentation method, \eg, Mask3D, with a 3D instance mask segmenter guided by 2D methods (\ie, merging superpoints with the help of a 2D segmenter, such as SAM). For each mask, Open3DIS computes a text-aligned representation by aggregating the CLIP visual representation of multi-scale crops from multiple views.

Our work focuses on 3D instance segmentation. However, unlike previous work on open-vocabulary 3D instance segmentation, \taskname features a novel setting that operates under the assumption that target classes are unknown during inference. \ourmethod addresses \taskname in a training-free manner by leveraging \vl assistants to provide scene semantics and ground them into 3D instance segments.

\section{Vocabulary-free 3D scene understanding}\label{sec:voc3d}
\vspace{-1mm}

\noindent\textbf{Definition.}
Vocabulary-free 3D instance segmentation (\taskname) aims to assign a semantic label to each 3D instance mask in a point cloud without relying on any predefined list of categories (vocabulary) at test time. 
Formally, given a point cloud $\mathcal{P} = \{\mathbf{p}\}$, where $\mathbf{p} \in \mathbb{R}^d$ s.t.~$d \ge 3$\footnote{
$d=3$ represents the basic case in which a point is represented by a 3D coordinate (LiDAR capture). 
In some instances, 
$d=4$ or $d=6$ if LiDAR luminance, or RGB information are available, respectively.},
The point cloud is decomposed into a set of 3D instance masks $\mathcal{M}^{3D}$, where each mask $M^{3D}_i \in \mathcal{M}^{3D}$ is a set of binary values with ones indicating its corresponding points belonging to the $i^{th}$ object instance, and zeros otherwise.  
\taskname involves assigning a class $c \in \mathcal{S}$ to each 3D instance mask $M^{3D}_i$, where $\mathcal{S}$ represents an unconstrained semantic space. For example, BabelNet~\cite{navigli2010babelnet} contains millions of semantic concepts, that is four magnitudes larger than the semantic classes annotated in ScanNet200~\cite{scannet200}. 
The objective is to design a function $f$ that maps 3D masks to concepts, formally defined as $f: \mathcal{M}^{3D} \rightarrow \mathcal{S}$.
At test time, the function $f$ has access to the point cloud $\mathcal{P}$ and to a source that provides vast semantic concepts approximating $\mathcal{S}$. Potential semantic sources, as discussed in~\cite{conti2023vocabularyfree}, can be either in the format of a web-scale database or a model that is trained with such database.
\taskname requires searching for relevant concepts from a vast semantic source and requires point-level semantic grounding, making it significantly more challenging than image classification as in \cite{conti2023vocabularyfree}.

\noindent\textbf{Challenges.}
A key challenge in distinguishing concepts within a vast semantic space in point clouds is ensuring spatial consistency when assigning labels to points. 
This requires models to not only label individual points accurately but also to guarantee that adjacent points, likely belonging to the same object, are assigned to the same label. 
Moreover, point sparsity may result in incomplete or ambiguous object representations, making it difficult to distinguish small or fine-grained objects. Additionally, reconstruction noise can render certain elements of the scene geometrically indistinguishable, making it challenging to differentiate between foreground and background or objects with similar shapes but different functions, particularly in cases where photometric information is missing or inaccurate.

\section{Our approach}\label{sec:approach}
\vspace{-1mm}
\begin{figure*}[t]
    \centering
    \includegraphics[width=\linewidth]{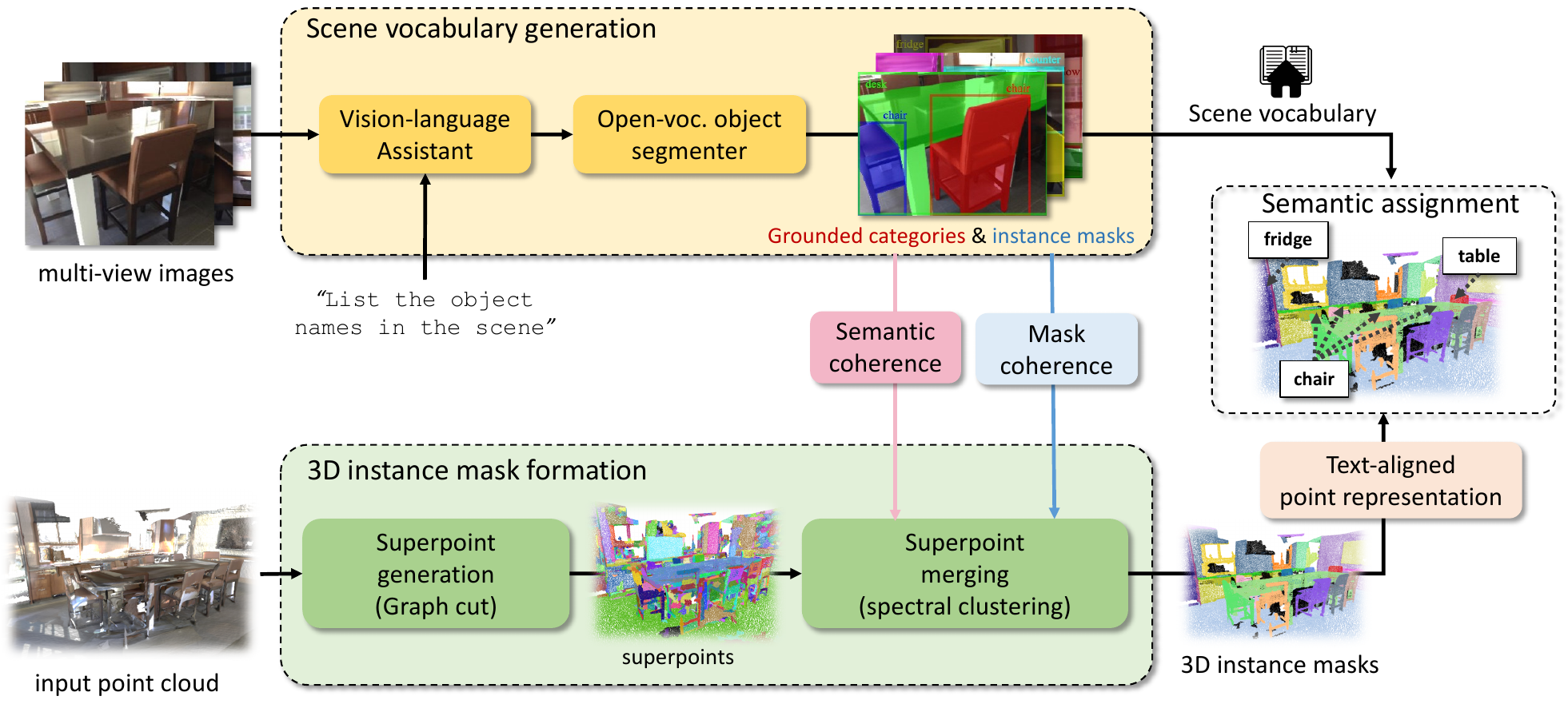}
    \vspace{-6mm}
    \caption{\ourmethod's architecture to address the \taskname task. To generate the scene vocabulary, we start from multi-view images and use a \vl assistant to identify lists of objects contained in each posed image. Then, we run an open-vocabulary object segmenter to ground the identified categories with instance masks to mitigate the potential risk of hallucination. We finally obtain the scene vocabulary, \ie, the list of unique grounded categories among all posed images.
    In parallel, to form 3D instance masks, we extract superpoints from the point cloud with graph cut. We then merge those dense superpoints to form 3D instance masks, considering both the \textit{semantic coherence} and \textit{mask coherence} which are computed with the 2D object masks. Finally, we obtain text-aligned point features for all points within each 3D instance mask, which are then used to assign the semantic category within the scene vocabulary. In addition to \taskname, \ourmethod is also able to deal with the open-vocabulary setting, by substituting the scene vocabulary with any predefined vocabulary or user-specified prompt.}
    \label{fig:architecture}
        \vspace{-5mm}
\end{figure*}

Given the point cloud $\mathcal{P}$ of a 3D scene and the corresponding set of $N$ posed images $\mathcal{V} = \{I_n\}_{n=1}^N$, the proposed \ourmethod predicts 3D instance masks with their associated semantic labels without knowing a predefined vocabulary.
As shown in \cref{fig:architecture}, \ourmethod first utilizes a large \vl assistant and an open-vocabulary 2D instance segmentation model to identify and ground objects on each posed image $I_n$, forming the scene vocabulary $\mathcal{C}$. 

Meanwhile, we partition the 3D scene $\mathcal{P}$ into geometrically-coherent superpoints $\mathcal{Q}$, to serve as initial seeds for 3D instance proposals.
Then, with the semantic-aware instance masks from multi-view images, we propose a novel procedure in representing superpoints and guiding their merging into 3D instance masks, using both the grounded semantic labels and their instance masks.
Specifically, by projecting each 3D superpoint onto image planes, and checking its overlapping with 2D instance masks, we can aggregate the semantic labels from multiple views within each superpoint. Once each superpoint is associated to a semantic label, we then perform superpoint merging to form 3D instance masks via spectral clustering. To do so, we define an affinity matrix among superpoints constructed by both mask coherence scores computed with the 2D instance masks, and semantic coherence scores computed with the per-superpoint textual embeddings.  
Finally, for each 3D instance proposal, we obtain the text-aligned representation by aggregating the CLIP visual representation of multi-scale object crops from multi-view images (as in~\cite{nguyen2024open3dis}). We further enrich such vision-based representation with textual representation coming from the merged superpoints. The text-aligned mask representation enables the semantic assignment to instance masks with the scene vocabulary~$\mathcal{C}$. 

It is worth to mention that \ourmethod can not only address \taskname, but it is also compatible with the open-vocabulary setting, by performing semantic assignment with any given vocabulary or user-specified prompt.

\subsection{Scene vocabulary generation} \label{sec:method:candidatevoc}
\ourmethod first utilizes a large \vl assistant to identify the scene vocabulary $\mathcal{C}$, \ie, the list of object categories in the scene, that are grounded in the multi-view images. 
Specifically, for each posed image $I_n$, we prompt the \vl assistant with \prompt. We then parse the response to obtain the list of objects, $\mathcal{C}_n^{-}$, present in the image from the answer. To mitigate the potential hallucination of object presence by the \vl assistant, we subsequently employ an open-vocabulary 2D instance segmentation model, \eg grounded SAM, to ground all categories in $\mathcal{C}_n^{-}$, obtaining the grounded object categories $\mathcal{C}_n$, as well as the set of masks $\mathcal{M}^{2D}_n$ for each object on the 2D posed image $I_n$. 

With $\mathcal{C}_n$, we then construct the scene vocabulary $\mathcal{C}$ for each point cloud by retaining only the unique categories from the combined sets of $ \mathcal{C}_{n} $, formally as $\mathcal{C} = \bigcup_{n=1}^{N} \mathcal{C}_{n}$,
where $\bigcup$ denotes the union operation.

The grounded object categories $\mathcal{C}_n$, and their corresponding instance masks $\mathcal{M}^{2D}_n$ on each 2D posed image, are further exploited in the process of representing and merging superpoints towards 3D instance masks $\mathcal{M}^{3D}$, as detailed in the following section. 

\subsection{3D instance mask formation}~\label{subsec:mask} 
We leverage geometrically coherent over-segmented superpoints to initialize 3D instance formation.
In addition to being a common practice in prior works~\cite{tai2024open,yin2024sai3d,nguyen2024open3dis}, superpoint initialization is more generalizable and better suited for zero-shot setups compared to pre-trained class-agnostic 3D instance segmentation models, such as Mask3D~\cite{schult2023mask3d}, which are trained on datasets used for method evaluation.
In the following, we detail the process of generating superpoints and how they are merged into 3D instances by leveraging the results of instance segmentation on posed images.

\noindent\textbf{Superpoint generation.} We use graph cut to group points into geometrically homogeneous regions, yielding a set of $M$ superpoints $\mathcal{Q} = \{Q_i\}_{i=1}^M$, where $Q_i$ is a binary mask of points in $\mathcal{P}$. Superpoints are dense partitions of the 3D scene. Neighboring superpoints are likely corresponding to the same semantic label. For example, a table, according to geometric features, might be partitioned into multiple superpoints corresponding to different surface planes, with each plane sharing the same semantic label, \ie, ``Table''. 
Via merging superpoints, we can form semantically-coherent 3D instance masks. 

\noindent\textbf{Superpoint merging by spectral clustering.} 
We aim to merge superpoints that tend to overlap with the same 2D masks when projected onto their respective images while ensuring semantic consistency. To this end, we define:
i) a \textit{mask coherence} score $a^M_{i,j}$ that quantifies the likelihood that two superpoints, $Q_i$ and $Q_j$, belong to the same object instance;
ii) a \textit{semantic coherence} score $a^S_{i,j}$ that quantifies the probability that two superpoints correspond to the same class;
iii) a \textit{spatial connectivity} score $a^C_{i,j}$ that indicates adjacency if the distance between the superpoints falls within a predefined threshold.

We derive the mask coherence score by evaluating the overlap ratio between the 3D superpoints $Q_i$ and $Q_j$ when projected onto the image plane of a 2D mask.
Specifically, for each 2D instance mask $M^{2D}_t {\in} {\{\mathcal{M}^{2D}_n\}}_{n=1}^{N}$, we calculate the Intersection over Union (IoU) $O_{i,t}$ between each superpoint $Q_i$ and the image pixels corresponding to $Q_i$. This is done by projecting all points of $Q_i$ onto the image plane of $M^{2D}_t$ using the known camera matrix, while excluding points outside the camera’s field of view. A superpoint is considered to have sufficient overlap with a 2D mask if the IoU is higher than a threshold, \ie, $O_{i,t} {>} \tau_{iou}$.
We then compute the mask coherence score $a^M_{ij}$ between two superpoints $Q_i$ and $Q_j$, accounting all 2D instance masks as:
\vspace{-2mm}
\begin{equation}
\vspace{-1mm}
    a^M_{ij} = \sum_{t} g\left(O_{i,t},\tau_{iou}\right)\cdot g\left(O_{j,t},\tau_{iou}\right),
\vspace{-1mm}
\end{equation}
where $T$ is the total number of 2D instance masks and the function $ g(x, \tau) $ is defined as $    g(x, \tau) = x$ if $ x > \tau$, and 0 otherwise.
By computing $a^M_{ij}$ among all pairs of superpoints, we obtain the mask coherence matrix $A^{M}$.

To compute the semantic coherence score $a^S_{ij}$, we first obtain the semantic representation per superpoint. Specifically, for each superpoint $Q_i$, we identify the top $K$ 2D masks based on the IoU $O_{i,t}$ and their corresponding semantic labels.
The most frequent label among these 2D masks is assigned as the semantic label for the superpoint $Q_i$. We then obtain its semantic representation by encoding the label text of $Q_i$ into a feature vector $f_{Q_i}$ using a text encoder.
We then calculate the cosine similarity using the semantic representation between $Q_i$ and $Q_j$ as $a^S_{ij}$:
\vspace{-2mm}
\begin{equation}
\vspace{-1mm}
    a^S_{ij} = \frac{f_{Q_i}^\top f_{Q_j}}{\|f_{Q_i}\|\|f_{Q_j}\|} \odot \left(f_{Q_i}^\top f_{Q_j} > \tau_{sim}\right).
\vspace{-1mm}
\end{equation}
Similarly, we obtain the semantic coherence matrix $A^{S}$ by computing $a^S_{ij}$ among all pairs of superpoints.

To efficiently compute the spatial connectivity scores denoted as an  adjacency matrix $A^C$, we sample points $K {=} 64$ (using the farthest point sampling) from each superpoint and set the threshold as twice the average distance $\tau_c$ of the sampled points. The distance between two superpoints $ Q_i $ and $ Q_j $ is defined as $ d(Q_i, Q_j) = \min_{p \in Q_i, q \in Q_j} \|p - q\| $. The adjacency score $ a^C_{ij} $ is then assigned as $ a^C_{ij} {=} 1 $ if $ d(Q_i, Q_j) {<} \tau_c $ and $ a^C_{ij} = 0 $ otherwise.

With $A^M$, $A^S$, and $A^C$, we construct an affinity matrix $A$ for the superpoints $\mathcal{Q}$ satisfying $A {=} A^M \odot A^S \odot A^C$.
Next, we compute the Laplacian eigenvectors of $A$, where the Laplacian matrix is defined as $L {=} D^{-1/2} (D {-} A) D^{-1/2}$. We then merge superpoints using its eigenvectors: $\{ y_0, \ldots, y_{M-1} \} {=} \text{eigs}(L)$, with the corresponding eigenvalues $\{\lambda_0, \ldots, \lambda_{M-1}\}$ ranked in ascending order, where $\lambda_0=0$.
We discretize the first $H$ eigenvectors $\{ y_0, \ldots, y_{H} \}$ of $L$ by clustering them across the eigenvector dimension using K-means. $H$ is determined by eigengap heuristic as $H {=} \argmax\limits_{1\leq j\leq M-2} \left(\lambda_{j+1}-\lambda_j\right).$
In doing so, we can extend $Q_i$ with neighboring superpoints $Q_j$ that meet the overlapping condition ($\tau_{iou}$) with the highest semantic similarity. Please refer to Sec. A of the supplementary materials for more details.

\subsection{Text-aligned point representation}\label{sec:method:pointembedding}
\vspace{-1.5mm}
Unlike prior works~\cite{jatavallabhula2023conceptfusion,peng2023openscene}, which leverage only the visual encoder of pre-trained \vl models such as CLIP, we employ both the vision encoder and the text encoder for per-point representation to mitigate the potential modality gap observed in~\cite{conti2023vocabularyfree}.
Specifically, given a point $\mathbf{p}_l$ and its corresponding superpoint $Q_i$, we first use the vision feature extraction method provided by Open3DIS~\cite{nguyen2024open3dis} to extract the per-point feature $f^v_l$. The CLIP vision encoder is then used to encode image crops from multiple posed images at multiple scales, obtained by projecting their corresponding 3D masks onto the posed images.
Finally, we enrich $f^v_l$ with the superpoint-level feature $f_{Q_i}$, obtaining the final point-level feature $f_l$ via mean pooling.

\section{Experiments}\label{sec:exp}
\vspace{-1.5mm}
We evaluate \ourmethod on the 3D scene instance segmentation task in open-vocabulary and vocabulary-free settings.
We compare \ourmethod with state-of-the-art methods by using two common benchmark datasets. 
We present quantitative and qualitative results and ablation studies.

\noindent\textbf{Datasets.}
We quantitatively evaluate \ourmethod by using 3D scans of real scenes from the ScanNet200~\cite{scannet200} and Replica~\cite{straub2019replica} datasets. 
These datasets include both instance and semantic (vocabulary) annotations.
ScanNet200~\cite{scannet200} contains a validation set of 312 indoor scans with 200 object categories, which is significantly more than its predecessor, that is ScanNet~\cite{dai2017scannet}, which features 20 semantic classes only.
Replica~\cite{straub2019replica} consists of 8 evaluation scenes with 48 classes. 
In the supplementary material, we also present additional results obtained on the S3DIS dataset~\cite{armeni20163d} (Sec.~B).

\noindent\textbf{Performance metrics.}
Following the experimental setup of ScanNet~\cite{dai2017scannet}, we compute the score at average mask thresholds ranging from 50\% to 95\% in 5\% increments as the Average Precision (AP). 
We then compute the AP at specific mask overlap thresholds of 50\% and 25\% as $\text{AP}{50}$ and $\text{AP}_{25}$, respectively. 
For ScanNet200, we report results for different category groups such as $\text{AP}_{\text{head}}$, $\text{AP}_{\text{com}}$, and $\text{AP}_{\text{tail}}$~\cite{scannet200}.

In the open-vocabulary setting, evaluation is straightforward as the vocabulary provided by the underlying dataset can be used directly, however in \taskname the \vl assistant can label objects differently than those labeled by humans in the ground truth \cite{conti2023vocabularyfree}.
We mitigate this problem by using the BERT Similarity~\cite{Zhang2020} that can quantify the semantic relevance of the predicted point label in relation to the ground-truth label, as in \cite{Koch2024}.
The BERT Similarity is 1 when the similarity between predicted and ground-truth labels is highest. 
We use a stringent threshold $\tau_{bert}=0.8$ on this similarity to deem a predicted label correct.

\noindent\textbf{Baselines.}\label{sec:baselines}
We compare \ourmethod with state-of-the-art methods, including OpenScene~\cite{peng2023openscene}, OpenMask3D~\cite{takmaz2023openmask3d}, OVIR-3D~\cite{lu2023ovir}, SAM3D~\cite{yang2023sam3d}, SAI3D~\cite{yin2024sai3d}, OVSAM3D~\cite{tai2024open}, and Open3DIS~\cite{nguyen2024open3dis}.
OpenScene is adapted for instance segmentation using Mask3D \cite{schult2023mask3d} as in \cite{nguyen2024open3dis}: we name this version OpenScene$^\star$.
Since there are no existing scene understanding methods specifically designed for the \taskname setting (SAM mask + vocabulary-free semantics), we implemented a set of baselines using state-of-the-art methods.
We adapt the open-vocabulary instance scene segmentation methods Open3DIS~\cite{nguyen2024open3dis} and SAM3D~\cite{yang2023sam3d} to the \taskname setting, and name these versions Open3DIS$^\dagger$ and SAM3D$^\dagger$, respectively.
We only use the 2D mask proposals provided by Open3DIS to be comparable with our method.
We replace the user-provided vocabulary, \ie, the full category list of each dataset, with the one generated by LLaVA as described in Sect.~\ref{sec:method:candidatevoc}.
Lastly, we evaluate \ourmethod in the open-vocabulary setting for further comparison of \ourmethod with state-of-the-art methods, \ie, 2D/3D mask + Open-vocab.~semantic.

\noindent\textbf{Implementation Details.}
\ourmethod is implemented with PyTorch using the original implementations of CLIP \cite{Radford2021clip}, \llava \cite{liu2023llava}, and Grounded-SAM\footnote{https://github.com/IDEA-Research/Grounded-Segment-Anything}.
For \llava, we use llava-v1.6-mistral-7b, while for CLIP, we use ViT-L/14. 
We set $\tau_{iou}=0.9, \tau_{sim}=0.9$ for all experiments.
For each superpoint, we choose top $K=5$ view masks with the largest IoU of projected
points.
Experiments are run on a single NVIDIA A40 48GB RAM.
We use the original source codes for the baselines.

\begin{table*}[t]
\centering
\vspace{-1mm}
\tabcolsep 12pt
\caption{3D instance segmentation results on ScanNet200. 
Best result for each metric is in \textbf{bold}.
}
\label{tab:performance_comparison}
\vspace{-4mm}
\resizebox{.88\textwidth}{!}{%
\begin{tabular}{lclccccc}
\toprule
\textbf{Method} & \textbf{Semantic} & \textbf{AP} & \textbf{AP$_{50}$} & \textbf{AP$_{25}$} & \textbf{AP$_{head}$} & \textbf{AP$_{com}$} & \textbf{AP$_{tail}$} \\
\midrule
\multicolumn{8}{c}{\cellcolor{SupMet}\textbf{3D mask + Open-vocab. semantic}} \\
\midrule
OpenScene$^*$ \cite{peng2023openscene} & OpenSeg \cite{ghiasi2022openseg} & {11.7} & {15.2} & {17.8} & {13.4} & {11.6} & {9.9} \\
OpenMask3D \cite{takmaz2023openmask3d} &  CLIP \cite{Radford2021clip} & 15.4 & 19.9 & 23.1 & 17.1 & 14.1 & 14.9 \\
\midrule
\multicolumn{8}{c}{\cellcolor{OpenVoc}\textbf{2D mask + Open-vocab. semantic}} \\
\midrule
OVIR-3D~\cite{lu2023ovir} &  Detic \cite{zhou2022detecting} & 9.3 & 18.7 & 25.0 & 9.8 & 9.4 & 8.5 \\
SAM3D \cite{yang2023sam3d} &  OpenSeg \cite{ghiasi2022openseg} & 7.4 & 11.2 & 16.2 & 6.7 & 8.0 & 7.6 \\
SAI3D \cite{yin2024sai3d} &  OpenSeg \cite{ghiasi2022openseg} & {9.6} & {14.7} & {19.0} & {9.2} & {10.5} & {9.1} \\
OVSAM3D~\cite{tai2024open} &  CLIP \cite{Radford2021clip} & {9.0} & {13.6} & {19.4} & {9.1} & {7.5} & {10.8} \\
Open3DIS \cite{nguyen2024open3dis} & CLIP \cite{Radford2021clip} & 18.2 & 26.1 & 31.4 & 18.9 & 16.5 & 19.2\\
\ourmethod &  CLIP \cite{Radford2021clip} & \bf22.4 & \bf27.9 & \bf34.4 & \bf20.7 & \bf20.2 & \bf20.6\\
\midrule
\multicolumn{8}{c}{\cellcolor{VocFree}\textbf{2D mask + Vocab.-free semantic}} \\
\midrule
SAM3D$^\dagger$~\cite{yang2023sam3d} &  OpenSeg \cite{ghiasi2022openseg} &  6.7 & 10.4 & 15.7 & 6.2 & 7.4 & 6.8 \\
Open3DIS$^\dagger$~\cite{nguyen2024open3dis} & CLIP \cite{Radford2021clip} & 17.6 & 25.4 & 30.9 & 18.4 & 15.8 & 18.2 \\
\ourmethod &  CLIP \cite{Radford2021clip} & \bf21.6 & \bf26.7 & \bf33.0 & \bf19.5 & \bf19.1 & \bf19.4 \\
\bottomrule
\end{tabular}
}
\vspace{-4mm}
\end{table*}

\subsection{Analysis of the results}
\vspace{-1.5mm}

\begin{figure*}[t]
\vspace{-3mm}
\centering
    \begin{tabular}{cccc}
        \multicolumn{2}{c}{
            \begin{overpic}[width=0.47\textwidth]{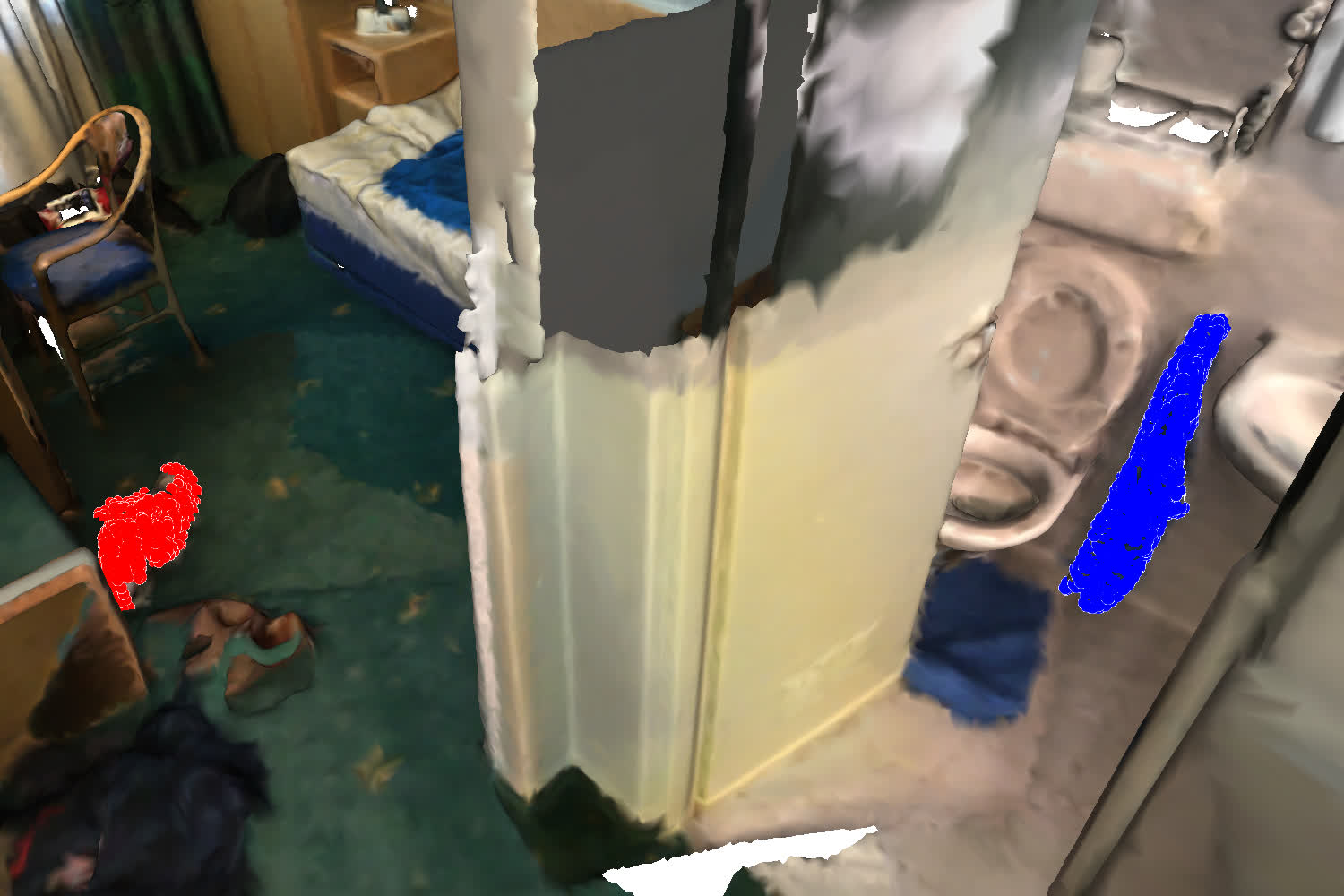}
            \put(10,20){\colorbox{White}{\footnotesize \textbf{running shoes}}}
            \put(75,20){\colorbox{White}{\footnotesize \textbf{blue towel}}}
            \end{overpic}
        }        
        & 
        \multicolumn{2}{c}{
            \begin{overpic}[width=0.47\textwidth]{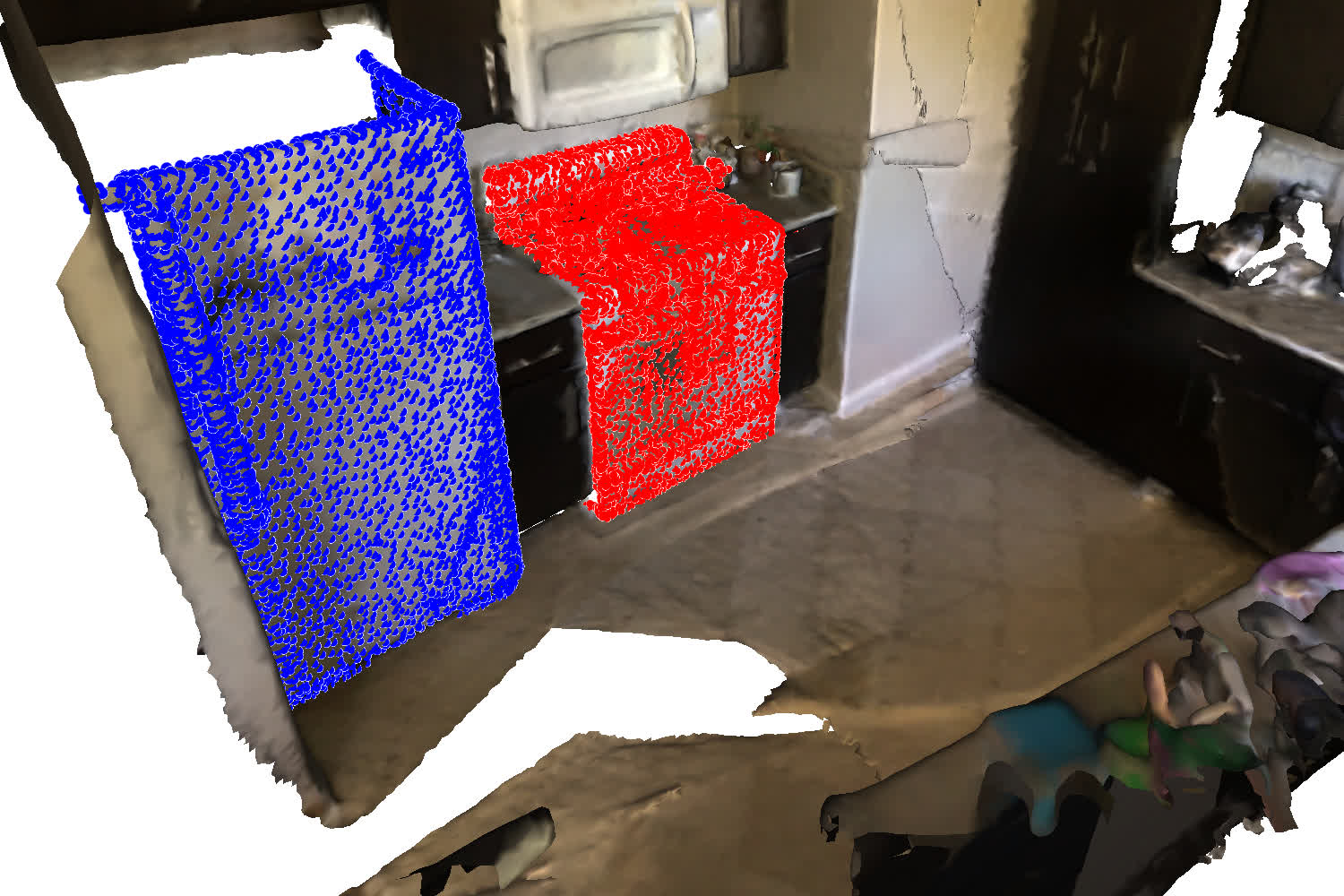}
            \put(15,25){\colorbox{White}{\footnotesize \textbf{chill}}}
            \put(45,30){\colorbox{White}{\footnotesize \textbf{gas plate}}}
            \end{overpic}
        }
        \\
        \begin{overpic}[width=0.22\textwidth]{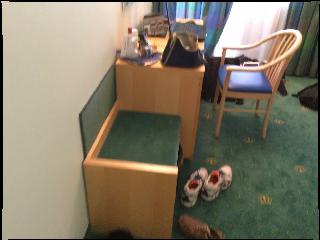}
        \put(52,8){\color{green}\dashbox{3}(25,20){}} 
        \end{overpic}
        &
        \begin{overpic}[width=0.22\textwidth]{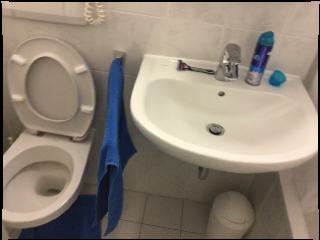}
        \put(30,1){\color{green}\dashbox{3}(15,60){}} 
        \end{overpic}
        &
        \begin{overpic}[width=0.22\textwidth]{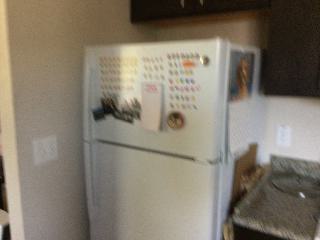}
        \put(25,1){\color{green}\dashbox{3}(60,64){}} 
        \end{overpic}
        &
        \begin{overpic}[width=0.22\textwidth]{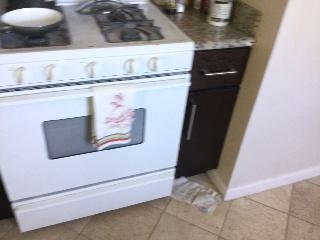}
        \put(0,1){\color{green}\dashbox{3}(61,70){}} 
        \end{overpic}
        \\
    \end{tabular}
    \vspace{-3mm}
    \caption{Qualitative results obtained by \ourmethod in the \taskname setting on ScanNet200. Instance masks are generated by querying \ourmethod with query vocabulary. The instance with the highest similarity score to the query's embedding is highlighted in the point clouds. Green boxes outline the regions of the objects in the corresponding RGB images.}
    \label{fig:qualitatives}
        \vspace{-4mm}
\end{figure*}

\noindent\textbf{ScanNet200.}
Tab.~\ref{tab:performance_comparison} reports \ovname and \taskname results on the ScanNet200 dataset. 
Following~\cite{hou2023mask3d,navigli2010babelnet}, we test both the \ovname and \taskname setting in the validation set.
The first and second sections of the table compare \ourmethod adapted to the open-vocabulary setting (3D/2D mask Open-vocab.~semantic) and baselines.
Although not explicitly designed for an open vocabulary setting, \ourmethod outperforms the other baselines.
A significant distinction between \ourmethod and Open3DIS lies in the processing of multi-view images. \ourmethod retrieves concepts through an assistant and transfers their features to the 3D points, whereas Open3DIS pre-processes images using open-vocabulary segmentation and transfers visual foundational features to the 3D points. 
We observed that the segmentation performance of Open3DIS is limited by this preprocessing and that the resulting image segmentation is relatively noisy.
In contrast, our method transfers and aggregates concepts based on vision and text features, followed by superpoint-based pooling to mitigate noisy features, thereby enhancing robustness.
The third section of the table reports the results in the \taskname setting, where we compare \ourmethod with Open3DIS$^\dagger$ and SAM3D$^\dagger$.
When Open3DIS$^\dagger$ is provided with the list of objects identified by LLaVA, its performance is inferior to that in the open-vocabulary setting. This suggests that Open3DIS$^\dagger$ struggles to handle a large corpus of concepts since it only considers vision features.
By contrast, \ourmethod significantly outperforms these baselines.
Specifically, \ourmethod achieves strong and stable performance across both common and rare class categories, as shown by the metrics AP$_{head}$, AP$_{com}$, and AP$_{tail}$. 
This is because our method is not trained on 3D annotated data, but instead leverages transferred semantic information from large vision-language models and 2D foundational models, which have been trained on massive 2D datasets and exposed to a wide range of categories.
Fig.~\ref{fig:qualitatives} shows the qualitative results of text-driven 3D instance segmentation. In the first row, we observe that \ourmethod can accurately segment most parts of the scene with the correct labels. 
Our model can successfully segment instances based on various types of input text prompts, which include object categories not present in the predefined labels, objects' functionality, branch, and other properties.
In the second row, we have highlighted the objects in the corresponding RGB images with boxes.
Sec.~B of the Supplementary Material analyses more results.

\noindent\textbf{Replica.}
Tab.~\ref{tab:replica} reports open-vocabulary and vocabulary-free results on the Replica dataset.
\ourmethod outperforms all the other baselines on both the open-vocabulary and \taskname settings.
Specifically, in the former, our approach outperforms Open3DIS~\cite{nguyen2024open3dis} and OVIR-3D~\cite{lu2023ovir} by margins of +2.7 and +9.7 in AP, respectively.
In the latter, \ourmethod outperforms Open3DIS$^\dagger$~\cite{nguyen2024open3dis} by a margin of +1.6 in AP.
This performance gap highlights the effectiveness of our approach in handling unseen categories, bolstered by the 2D foundation model and vision-language model assistance.

\begin{table}[t]
\centering
\vspace{-1mm}
\tabcolsep 12pt
\caption{3D instance segmentation results on Replica. 
Best result for each metric is in \textbf{bold}.
}
\label{tab:replica}
\vspace{-3mm}
\resizebox{.88\columnwidth}{!}{%
\begin{tabular}{lccc}
\hline
\textbf{Method} & \textbf{AP} & \textbf{AP$_{50}$} & \textbf{AP$_{25}$} \\ 
\midrule
\multicolumn{4}{c}{\cellcolor{SupMet}\textbf{3D mask + Open-vocab. semantic}} \\
\midrule
OpenScene$^\star$~\cite{peng2023openscene} & 10.9 & 15.6  & 17.3 \\
OpenMask3D~\cite{takmaz2023openmask3d} & 13.1 & 18.4 & 24.2 \\
\midrule
\multicolumn{4}{c}{\cellcolor{OpenVoc}\textbf{2D mask + Open-vocab. semantic}} \\
\midrule
OVIR-3D~\cite{lu2023ovir} & 11.1 & 20.5 & 27.5      \\
Open3DIS~\cite{nguyen2024open3dis}  & {18.1} & {26.7} & {30.5}        \\
\ourmethod & \bf20.8 & \bf28.7 & \bf34.4\\
\midrule
\multicolumn{4}{c}{\cellcolor{VocFree}\textbf{2D mask + Vocab.-free semantic}} \\
\midrule
Open3DIS\textsuperscript{\textdagger}~\cite{nguyen2024open3dis} & 17.3 & 25.8 & 29.0    \\
\ourmethod & \bf18.9 & \bf27.6 & \bf31.9   \\ 
\bottomrule
\end{tabular}
}
\vspace{-4mm}
\end{table}

\subsection{Ablation study}
\vspace{-1.5mm}
To evaluate the effectiveness of our model design, we conducted a series of ablation studies on the validation set of ScanNet200.
More ablation studies are given in the Supplementary Material.

\noindent\textbf{How effective are superpoints to guide mask representation?}
There are two key technical designs that we consider important for \ourmethod: text-embedding-enhanced features and superpoint-based average pooling.
Tab.~\ref{tab:ab_component} shows the performance of various feature fusion strategies. 
The first row displays the results using only vision features, denoted as VisEmb. We use the CLIP visual encoder to obtain visual representations for each 3D proposal by aggregating information from multiple views.
While VisEmb outperforms using only text representations (TxtEmb), their fusion enables \ourmethod to achieve overall and consistently improved results.
Since the generated point-level features may be noisy, we introduce a feature refinement module that averages representations (pooling) at the superpoint level, assuming that points belonging to the same superpoint should share the same features.
The third row (w/o SPool) shows the results without superpoint-based pooling, which are inferior to those of \ourmethod across all metrics.

\begin{table}[t]
    \centering
    \tabcolsep 5pt
    \caption{Ablation study of \ourmethod for instance feature extraction on ScanNet200 dataset.
    Each variant of \ourmethod has one component that differs from the final version of \ourmethod.
    }
    \vspace{-4mm}
    \label{tab:ab_component}
    \resizebox{\columnwidth}{!}{%
    \begin{tabular}{l c c c c c c}
        \toprule
        \bf Setting & \textbf{AP} & \textbf{AP$_{50}$} & \textbf{AP$_{25}$} & \textbf{AP$_{head}$} & \textbf{AP$_{com}$} & \textbf{AP$_{tail}$}\\
        \midrule
        VisEmb & 20.4 & 25.8 & 32.3 & 18.6 & 18.1 & 18.2 \\
        TxtEmb & 15.3 & 22.3 & 28.6 & 17.4 & 17.3 & 17.6 \\
        w/o SPool & 20.3 & 25.6 & 31.9 & 18.4 & 18.0 & 18.2  \\
        \ourmethod & \bf21.6 & \bf26.7 & \bf33.0 & \bf19.5 & \bf19.1 & \bf19.4 \\
        \bottomrule
    \end{tabular}
    }
    \vspace{-4mm}
\end{table}

\begin{table}[t]
    \centering
    \tabcolsep 5pt
    \caption{Ablation study of \ourmethod using text embedding for superpoint merging and superpoint-based pooling on ScanNet200.}
    \label{tab:ab_mask}
    \vspace{-3mm}
    \label{tab:ab_merge}
    \resizebox{\columnwidth}{!}{%
    \begin{tabular}{l c c c c c c}
        \toprule
        \textbf{Setting} & \textbf{AP} & \textbf{AP$_{50}$} & \textbf{AP$_{25}$} & \textbf{AP$_{head}$} & \textbf{AP$_{com}$} & \textbf{AP$_{tail}$}\\
        \midrule
        w/o TxtSim & 19.8 & 25.1 & 30.9 & 17.7 & 17.3 & 17.2 \\
        \ourmethod & \bf21.6 & \bf26.7 & \bf33.0 & \bf19.5 & \bf19.1 & \bf19.4 \\
        \bottomrule
    \end{tabular}
    }
    \vspace{-5mm}
\end{table}
\noindent\textbf{How effective is text embedding superpoint merging?} 
We experimentally assess the influence of text feature similarity (TxtSim) guided merging on the performance of \ourmethod. 
Second row of Tab.~\ref{tab:ab_mask} shows that incorporating text embedding for superpoint merging can enhance instance segmentation performance.
This improvement is achieved because 2D masks can encompass background regions or nearby objects, rendering IoU alone insufficient for accurately determining the association of superpoints with a 3D proposal. 
Leveraging text feature similarity helps \ourmethod to mitigate this issue.
\section{Conclusions}
\vspace{-2mm}
We presented a novel approach to 3D scene understanding that operates without the need for a predefined vocabulary. By integrating a large \vl assistant with an open-vocabulary 2D instance segmenter, our \ourmethod can autonomously identify and label each 3D instance in a scene. Furthermore, our innovative use of superpoints, in conjunction with spectral clustering, enables our system to generate robust 3D instance proposals.
We evaluated \ourmethod on two point cloud datasets, ScanNet200 and Replica, and showed that \ourmethod outperforms recent approaches adapted to the \taskname setting, as well as in the open vocabulary setting.

Because \ourmethod can effectively exploit \vl assistant understanding with point cloud data in a training-free manner, an exciting future research direction includes exploring new 3D scene understanding tasks such as affordances, using the soon-to-be-released dataset SceneFun3D \cite{Delitzas2024SceneFun3D}. 
Moreover, our goal is to improve geometric understanding using the most recent zero-shot approaches, such as \cite{mei2023geometrically}. 
Lastly, implementing new large \vl assistants can be a viable way to enhance point cloud understanding even further.

\noindent\textbf{Acknowledgment.}~This work was sponsored by PNRR ICSC National Research Centre for HPC, Big Data and Quantum Computing (CN00000013) and the FAIR - Future AI Research (PE00000013), funded by NextGeneration EU.

{
    \small
    \bibliographystyle{ieeenat_fullname}
    \bibliography{main}
}
\clearpage
\setcounter{page}{1}
\maketitlesupplementary

\renewcommand{\thesection}{\Alph{section}}
\setcounter{section}{0}
We first provide more details of the superpoint merging procedure via spectral clustering. 
Then we provide additional comparative analysis on the S3DIS dataset, and we perform an ablation on the hyperparameters of \ourmethod.
Lastly, we present more qualitative results on ScanNet200 and Replica datasets and implementation details for \taskname setting.

\section{{Spectral clustering}}\label{sec:rationale}

To cluster the superpoints more efficiently, we propose a hierarchical spectral clustering algorithm to generate masks. First, a Hilbert curve is applied to serialize the superpoints based on the coordinates of the center point $\mathbf{q}_i = \frac{1}{N_i} \sum\limits_{v_{j} \in Q_i} \mathbf{p}_{j}$ for each superpoint \( Q_i \), where $\mathbf{p}_{j}$ is a point in \( Q_i \) and \( N_i \) is the number of points in \( Q_i \). 
Next, a sliding window is used to divide the serialized superpoints into \( K_s \) groups, with each group containing \( N_s \) superpoints. 
The window shifts by a stride of \( N_s \). Spectral clustering is then applied to each group to generate coarse masks. This is an iterative spectral clustering process, where masks are merged with those from neighboring groups based on the overlap scores between two coarse masks, \( M_s \) and \( M_t \). The overlap score between \( M_s \) and \( M_t \) is determined by selecting the maximum similarity (from the affinity matrix \( A \)) between superpoints, where one superpoint belongs to \( M_s \) and the other to \( M_t \). 
Clustering combines region sets from merged coarse masks until \( M_s * M_t \) is less than the threshod $\tau_{iou}*\tau_{sim}$.
In our implementation, we first divide the superpoints into two groups for each 3D scene, followed by an iterative spectral clustering process.

\section{Additional experiments}
\subsection{S3DIS dataset}
S3DIS~\cite{armeni20163d} consists of 271 scenes that cover 13 classes in 6 areas. 
We adopt the Open3DIS categorization strategy~\cite{nguyen2024open3dis} for S3DIS, which splits the dataset into base and novel sets. The novel set includes two parts \textbf{N4} and \textbf{N6}.
PLA~\cite{ding2023pla}, Lowis3D~\cite{ding2024lowis3d}, and Open3DIS~\cite{nguyen2024open3dis} train on the base classes of S3DIS data, whereas the novel classes are not seen during training.
Note that our method is zero-shot and does not require any training (neither base nor novel classes). 

Tab.~\ref{tab:supp_ov3dis_results} compares the performance of our method with other approaches on S3DIS, focusing on Average Precision (AP) at 50\%  (AP\(_{50}\)) IoU for novel classes. 
The table compares methods across two settings: open-vocabulary and vocabulary-free (\taskname) setting. 
In the open-vocabulary setting, \ourmethod achieves the highest scores despite using 2D masks, with N4 AP\(_{50}\) of 29.1 and N6 AP\(_{50}\) of 33.4, outperforming methods like Open3DIS, which records 26.3 and 29.0 for N4 AP$_{50}$ and N6 AP$_{50}$, respectively.
Also in the vocabulary-free setting \ourmethod outperforms the other methods, achieving N4 AP\(_{50}\) of 28.4 and N6 AP\(_{50}\) of 29.7. 
Although this performance is slightly lower than in the open-vocabulary setting, it outperforms Open3DIS, which achieves 24.6 and 26.3 for N4 AP\(_{50}\) and N6 AP\(_{50}\), respectively.
These results demonstrate that \ourmethod provides superior performance in both open-vocabulary and vocabulary-free settings, highlighting its robustness and effectiveness in various 3D instance segmentation scenarios.

\begin{table}[t]
\centering
\tabcolsep 7pt
\caption{OV-3DIS results on S3DIS in terms of $AP_{50}$.}
\label{tab:supp_ov3dis_results}
\vspace{-3mm}
\begin{tabular}{lccc}
\toprule
\textbf{Method} & Mask type & \textbf{N4 $AP_{50}$} & \textbf{N6 $AP_{50}$} \\
\midrule
\multicolumn{4}{c}{\cellcolor{SupMet}\textbf{Open-vocab. semantic}} \\
\midrule
PLA~\cite{ding2023pla} & 3D mask & 8.6 & 9.8 \\
Lowis3D~\cite{ding2024lowis3d} & 3D mask & 13.8 & 15.8 \\
Open3DIS~\cite{nguyen2024open3dis} & 3D mask & 26.3 & 29.0\\
\ourmethod~\cite{nguyen2024open3dis} & 2D mask & \textbf{29.1} & \textbf{33.4} \\
\midrule
\multicolumn{4}{c}{\cellcolor{VocFree}\textbf{2D mask + Vocab.-free semantic}} \\
\midrule
Open3DIS~\cite{nguyen2024open3dis} & 3D mask & 24.6 & 26.3 \\
\ourmethod~\cite{nguyen2024open3dis} & 2D mask & \textbf{28.4} & \textbf{29.7} \\
\bottomrule
\end{tabular}
\end{table}

\begin{figure}[t]
\centering
\tabcolsep 1pt
 \begin{tabular}{cc}
        \begin{overpic}[width=0.23\textwidth]{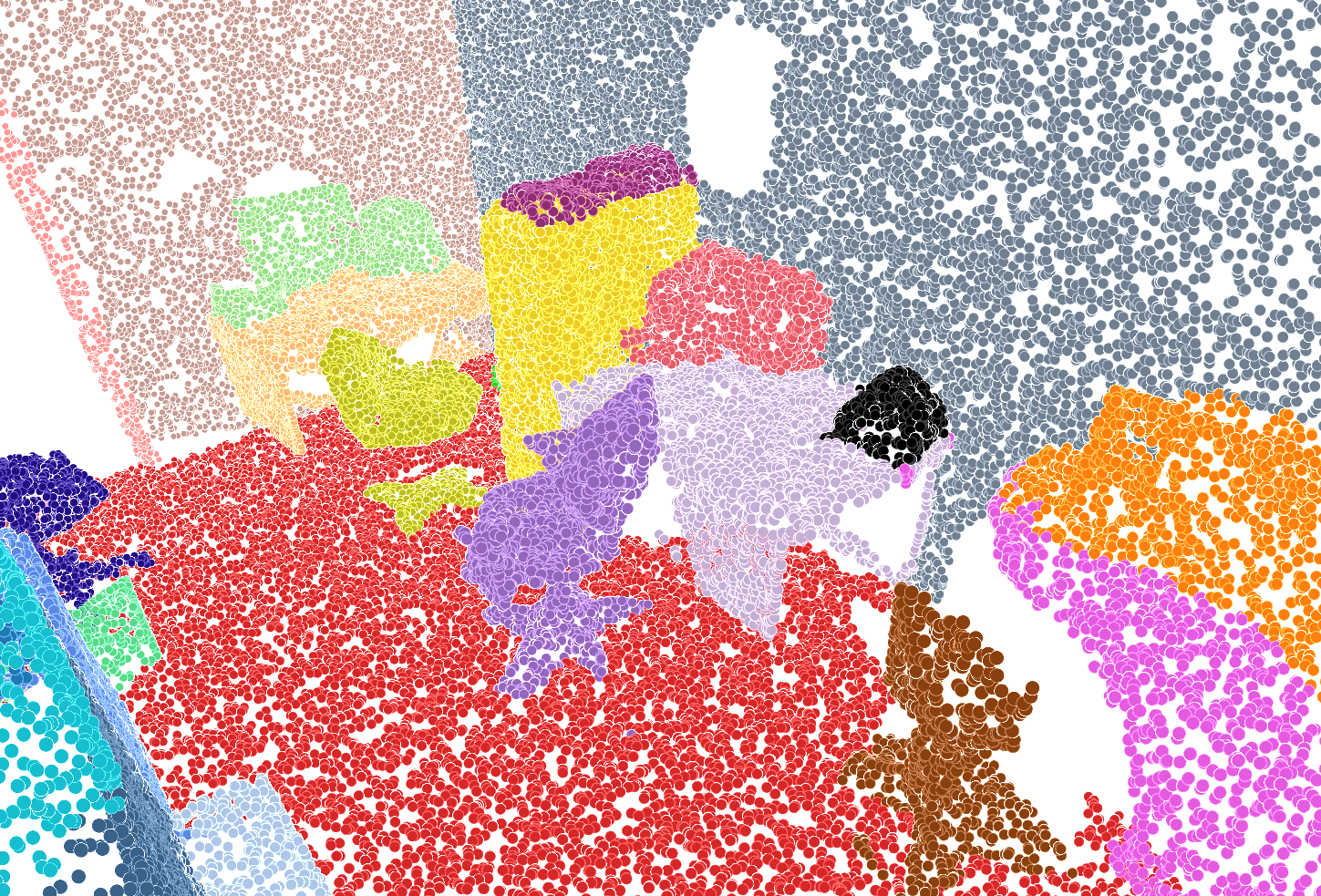}
            \put(32,72){{\footnotesize \textbf{Ins. GT}}}
            \end{overpic}
        &
        \begin{overpic}[width=0.23\textwidth]{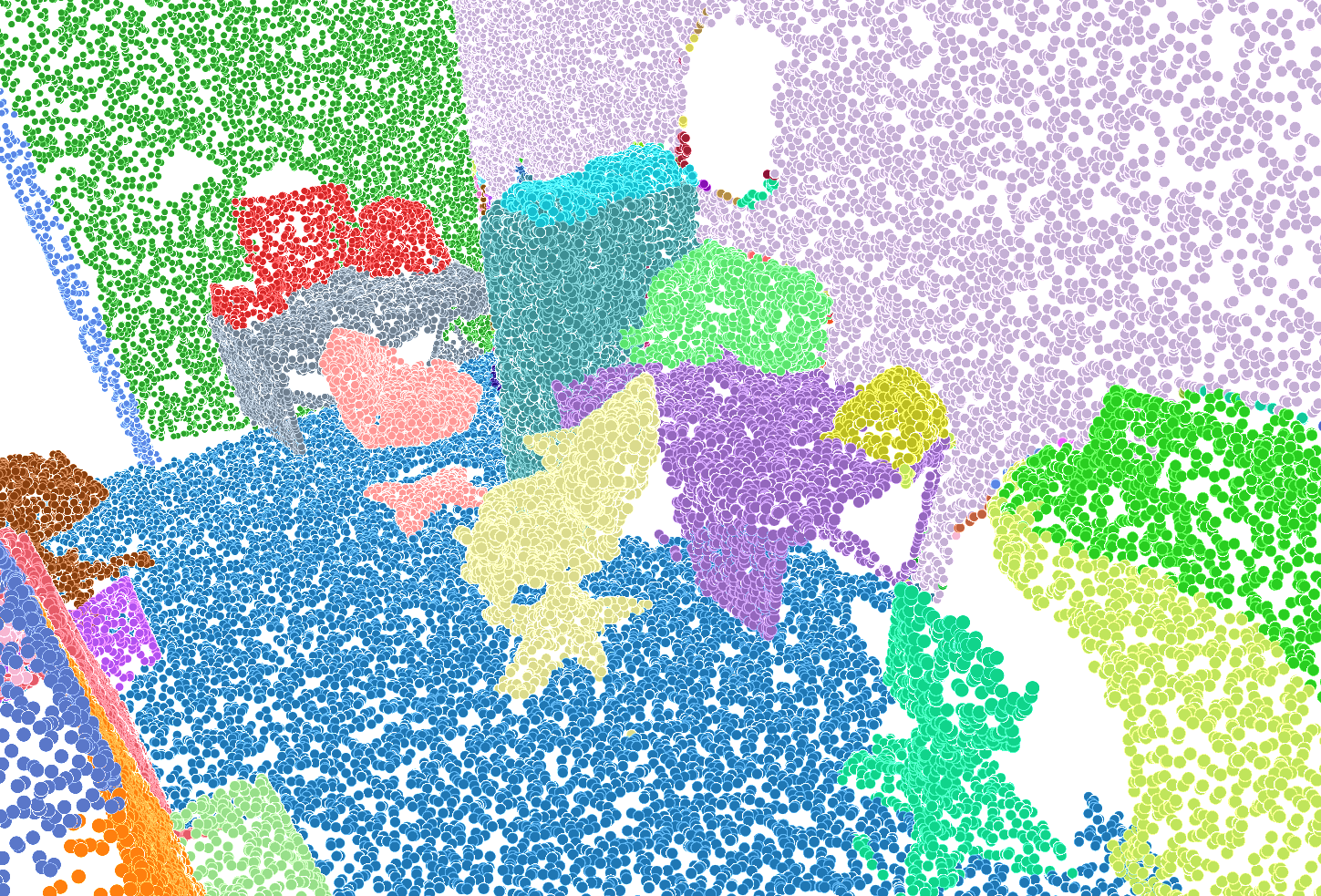}
            \put(32,72){{\footnotesize{\textbf{Ins. Pred.}}}}
            \end{overpic}
        \\
    \end{tabular}
    \vspace{-3mm}
    \caption{Qualitative results obtained by \ourmethod in the \taskname setting. Left to right: ground truth instance, predicted instance.
    }
    \label{fig:supp_s3dis}
\end{figure}

Fig.~\ref{fig:supp_s3dis} further presents two examples of instance segmentation results on S3DIS. Compared to the ground truth, our \ourmethod produces nearly identical results. This shows that our approach effectively leverages LLaVA-guided 2D mask prediction in conjunction with a superpoints-based spectral clustering strategy, resulting in high-quality instance segmentation results.

\subsection{Ablation study on different values of IoU and similarity threshold}
We use mask IoU and text similarity to guide 3D instance generation (superpoint merging). This involves two hyperparameters: the IoU threshold $\tau_{iou}$ and the similarity threshold $\tau_{sim}$, which determine how IoU and text similarity are considered when deciding whether two superpoints belong to the same 3D instance.
To evaluate the impact of the IoU threshold $\tau_{iou}$ and the similarity threshold $\tau_{sim}$ on the performance of instance segmentation, we report the performance of 3D instance mask formation from 2D masks extracted by Grounded-SAM from multi-view RGB images, using different values for the IoU threshold $\tau_{iou}$ and similarity threshold $\tau_{sim}$ in Tab.~\ref{tab:sup_iou} and Tab.~\ref{tab:sup_sim}. The tests were conducted on the ScanNet200 validation set.

\begin{table}[t]
\centering
\tabcolsep 8pt
\caption{Ablation study on IoU threshold (\(\tau_{iou}\)) and its impact on Average Precision (AP) and AP at 50\% IoU (AP\(_{50}\)).}
\label{tab:sup_iou}
\vspace{-3mm}
\begin{tabular}{c|ccccc}
\toprule
\(\tau_{iou}\) & 0.5 & 0.7 & 0.8 & 0.9 & 0.95 \\
\toprule
\multicolumn{6}{c}{\cellcolor{SupMet}\textbf{Open-vocabulary}} \\
\midrule
AP & 21.9 & 22.2 & 22.3 & \bf22.4 & 22.1\\
AP\(_{50}\) & 27.4 & 27.6 & \bf27.9 & \bf27.9 & 27.5 \\
\midrule
\multicolumn{6}{c}{\cellcolor{VocFree}\textbf{Vocabulary-free}} \\
\midrule
AP & 21.2 & 21.3 & 18.0 & \bf21.6 & 16.9 \\
AP\(_{50}\) &  26.2 & 26.4 & 26.6 & \bf26.7 & 26.4 \\
\bottomrule
\end{tabular}
\vspace{-3mm}
\end{table}

For the open-vocabulary setting, Tab.~\ref{tab:sup_iou} reports the results on \(\tau_{iou}\) quantified using average precision (AP), and AP at 50\% IoU (AP\(_{50}\)), by varying \(\tau_{iou}\) values from 0.5 to 0.95.
AP gradually increases from 21.9 at \(\tau_{iou} = 0.5\) to a peak of 22.4 at \(\tau_{iou} = 0.9\), before slightly decreasing at \(\tau_{iou} = 0.95\). AP\(_{50}\) follows a similar trend, peaking at 27.9 for both \(\tau_{iou} = 0.8\) and \(\tau_{iou} = 0.9\).
In the vocabulary-free setting, AP also shows an increase, reaching its highest value of 21.6 at \(\tau_{iou} = 0.9\) before dropping at \(\tau_{iou} = 0.95\). The AP\(_{50}\) increases from 26.2 at \(\tau_{iou} = 0.5\) to a maximum of 26.7 at \(\tau_{iou} = 0.9\), and then slightly decreases at \(\tau_{iou} = 0.95\).
These results indicate that using a higher \(\tau_{iou}\) threshold up to 0.9 generally improves performance in both settings. However, pushing the threshold to 0.95 can lead to performance degradation, suggesting that \(\tau_{iou} = 0.9\) provides the best results.

For the setting without vocabulary, Tab.~\ref{tab:sup_sim} reports the results of \(\tau_{\text{sim}}\) using AP, and AP\(_{50}\). By varying \(\tau_{\text{sim}}\) values from 0.5 to 0.9, both AP and AP\(_{50}\) improve, with AP peaking at 21.6 and AP\(_{50}\) reaching its highest value of 26.7 at \(\tau_{\text{sim}} = 0.9\). 
However, further increasing the threshold to 0.95 results in a slight decrease in both metrics, with AP dropping to 21.4 and AP\(_{50}\) to 26.3. These results suggest that while a higher similarity threshold generally enhances performance, setting the threshold too high can lead to lower performance, indicating that \(\tau_{\text{sim}} = 0.9\) offers the best results.

\begin{table}[t]
\centering
\tabcolsep 7pt
\caption{Ablation study on IoU threshold (\(\tau_{sim}\)) and its impact on Average Precision (AP) and AP at 50\% IoU (AP\(_{50}\)).}
\label{tab:sup_sim}
\vspace{-3mm}
\begin{tabular}{c|cccccc}
\toprule
\(\tau_{iou}\) & 0.5 & 0.6 & 0.7 & 0.8 & 0.9 & 0.95\\
\toprule
\multicolumn{7}{c}{\cellcolor{VocFree}\textbf{Vocabulary-free}} \\
\midrule
AP & 20.5 & 20.9 & 21.2 & 21.4 & \bf21.6 & 21.4\\
AP\(_{50}\) & 25.6 & 26.2 & 26.4 & 26.5 & \bf26.7 & 26.3\\
\bottomrule
\end{tabular}
\end{table}

\subsection{Visualizations on Replica}
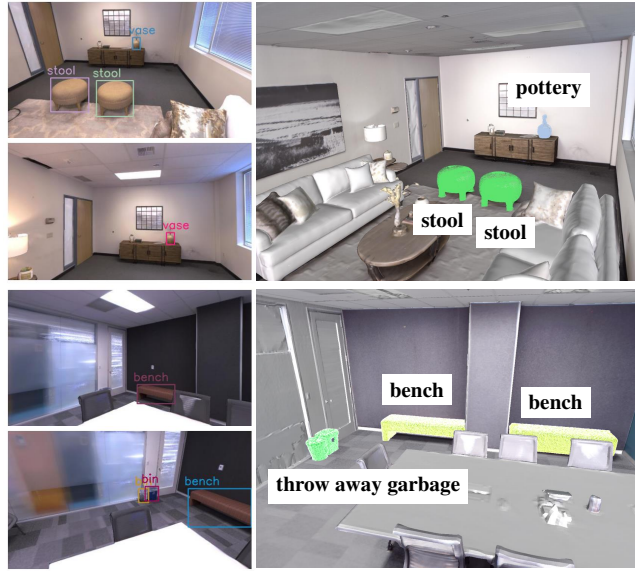
\begin{figure}[t]
\vspace{-3mm}
    \begin{overpic}[width=0.48\textwidth]{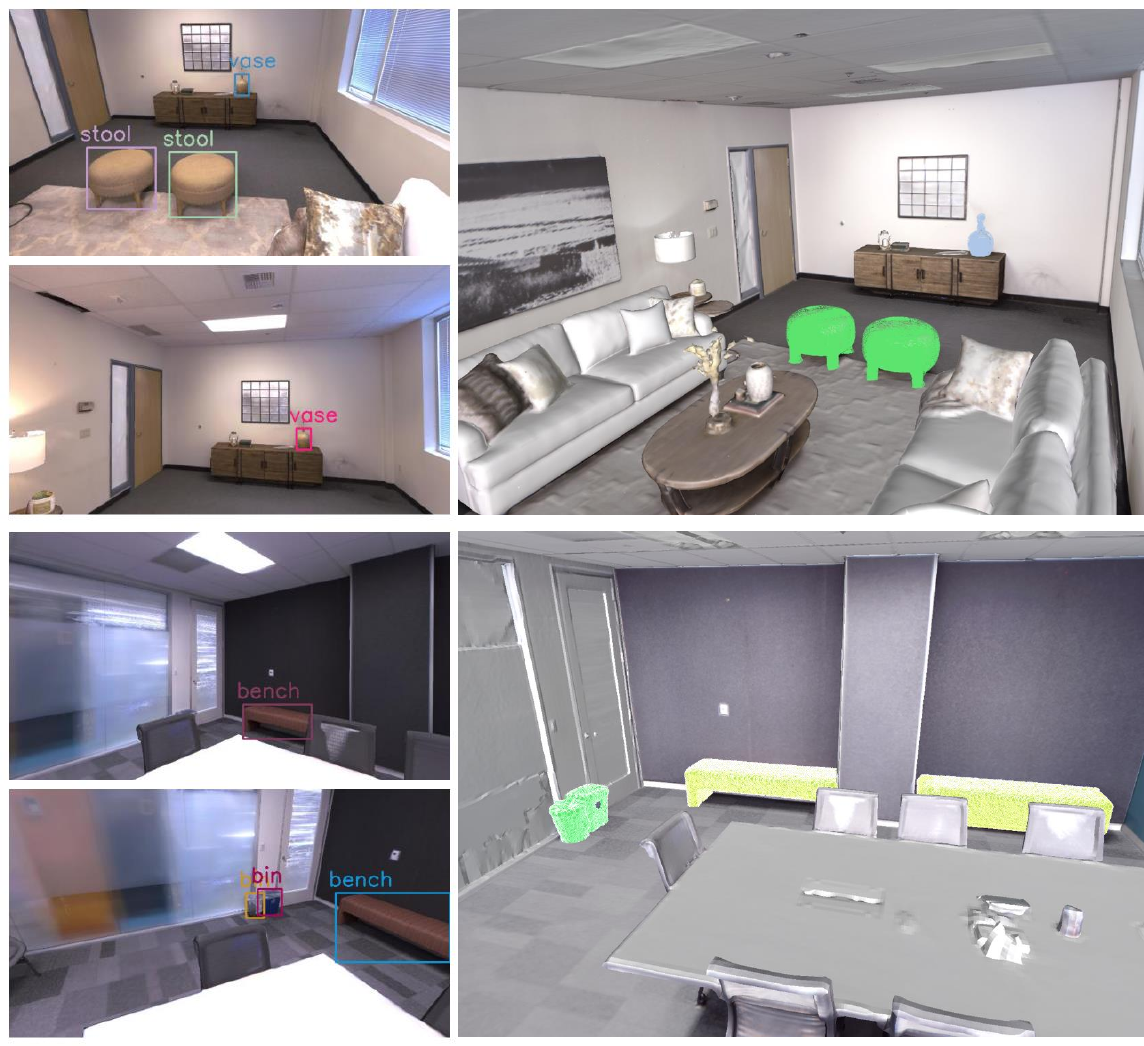}
    \put(65,55){\colorbox{White}{\footnotesize \textbf{stool}}}
    \put(75,53){\colorbox{White}{\footnotesize \textbf{stool}}}
    \put(80,76){\colorbox{White}{\footnotesize \textbf{pottery}}}
    \put(82,26){\colorbox{White}{\footnotesize \textbf{bench}}}
    \put(60,28){\colorbox{White}{\footnotesize \textbf{bench}}}
    \put(42,13){\colorbox{White}{\footnotesize \textbf{throw away garbage}}}
    \end{overpic}
    \vspace{-5mm}
    \centering
    \caption{Qualitative results of two examples obtained by \ourmethod in the \taskname setting on Replica dataset. The instance with the highest similarity score to the query's embedding is highlighted in the point clouds. In the images, each box outlines the regions of the objects detected by Grounded-SAM based on the queries.
    }
    
    \label{fig:supp_replica_query}
\end{figure}

Fig.~\ref{fig:supp_replica_query} shows the qualitative results of text-driven 3D instance segmentation. 
Our model can successfully segment instances based on various types of input text prompts, which can include object categories (such as pottery) not present in the predefined labels, or objects' functionality (throw away garbage).

\begin{figure*}[t]
    \centering
    \tabcolsep 1pt
    \begin{tabular}{cc|cc}
        \begin{overpic}[width=0.235\textwidth]{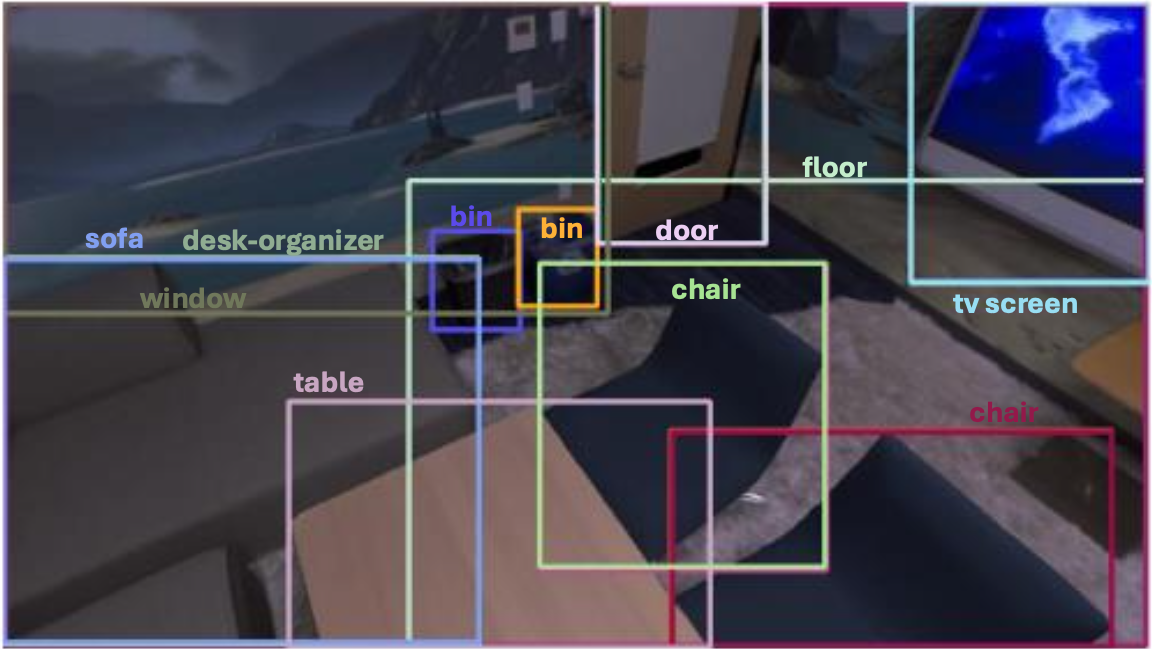}
            \put(78,60){\color{black}\textbf{Example (1)}}
            \put(-7,2){\color{black}\rotatebox{90}{\textbf{2D prediction}}}
        \end{overpic} &
        \begin{overpic}[width=0.235\textwidth]{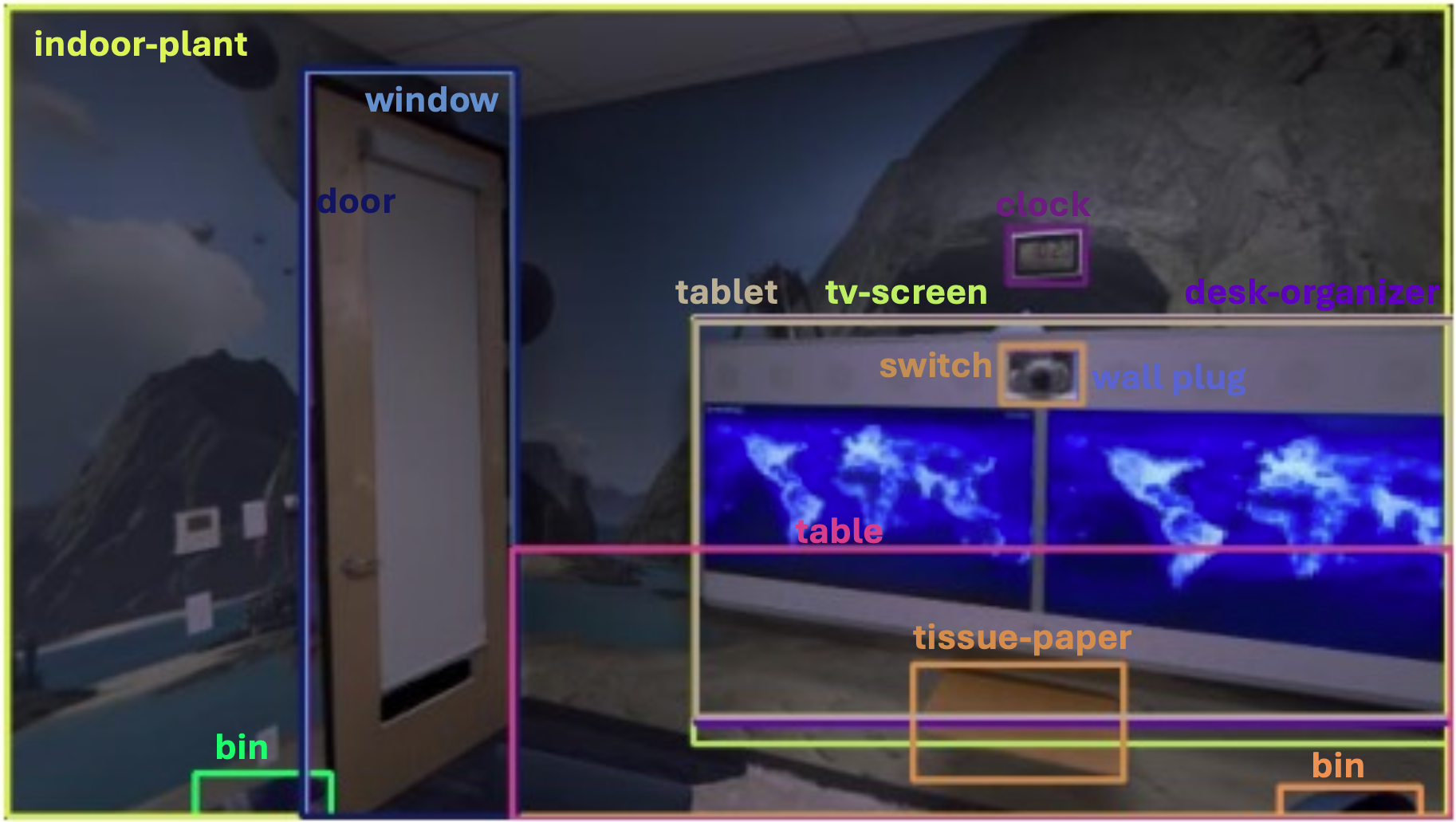}
        \end{overpic} &
        \begin{overpic}[width=0.235\textwidth]{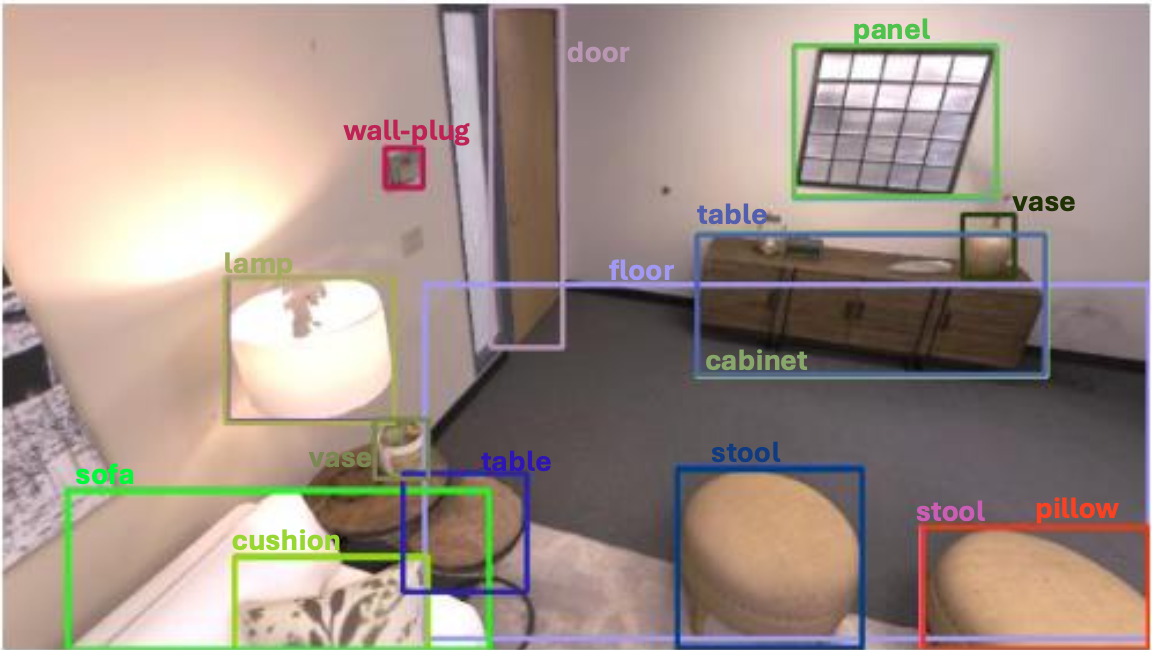}
            \put(78,60){\color{black}\textbf{Example (2)}}
        \end{overpic} &
        \begin{overpic}[width=0.235\textwidth]{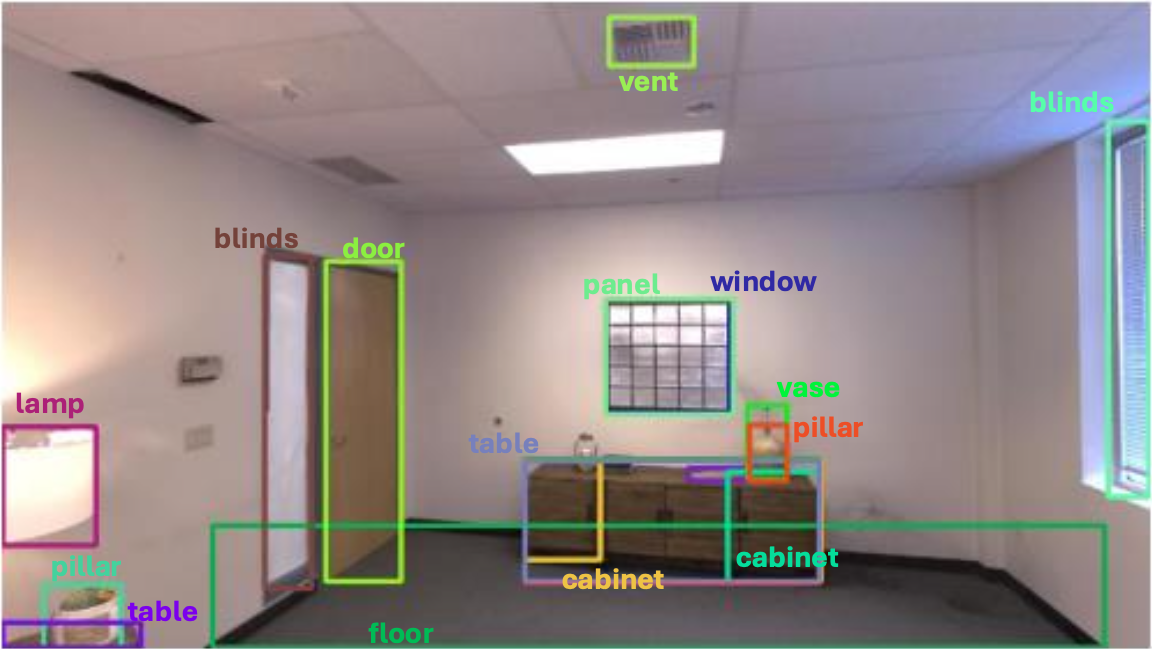}
        \end{overpic} \\
        \begin{overpic}[width=0.235\textwidth]{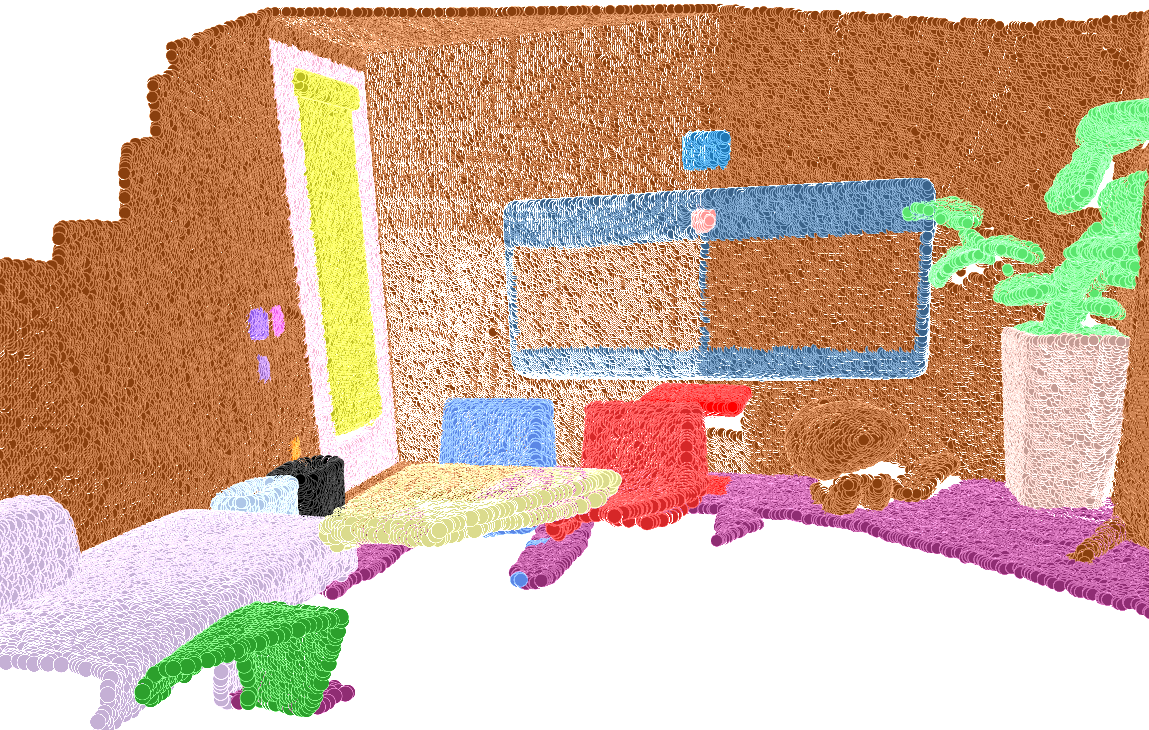}
                \put(40,-7){\color{black}\textbf{Ground-truth}}
                \put(-7,2){\color{black}\rotatebox{90}{\textbf{3D prediction}}}
            \end{overpic}
        &
        \begin{overpic}[width=0.235\textwidth]{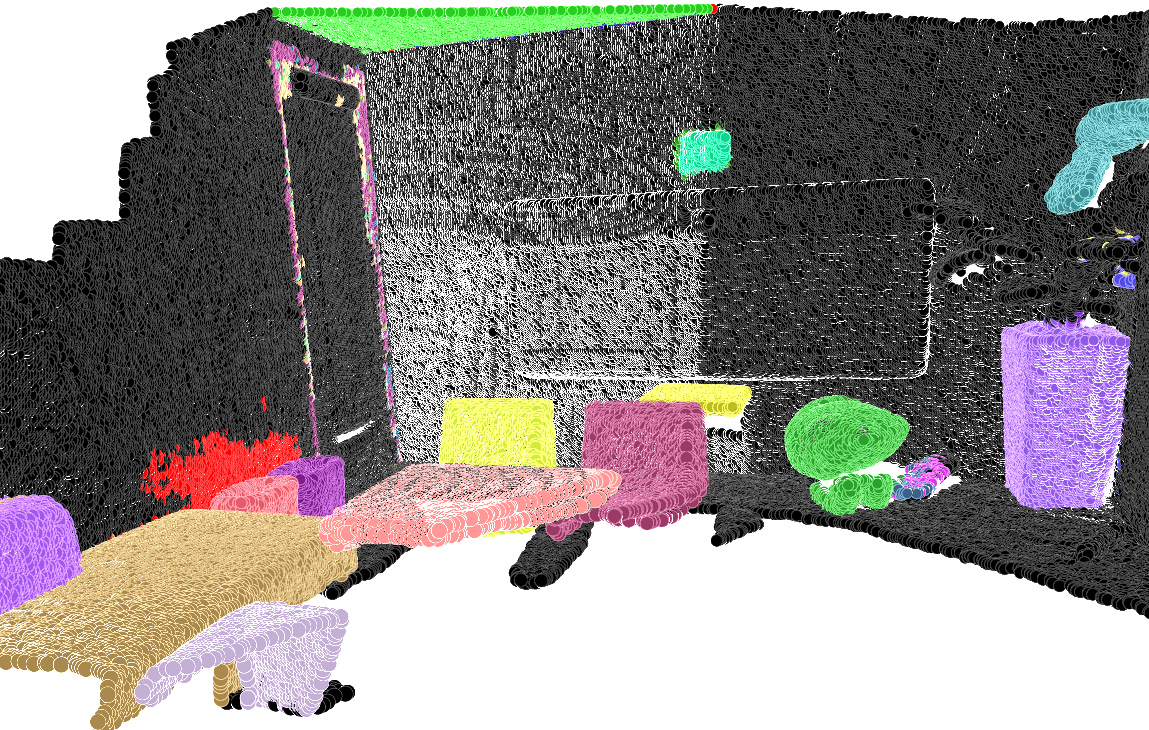}
            \put(40,-7){\color{black}\textbf{Prediction}}
        \end{overpic} 
        &
        \begin{overpic}[width=0.235\textwidth]{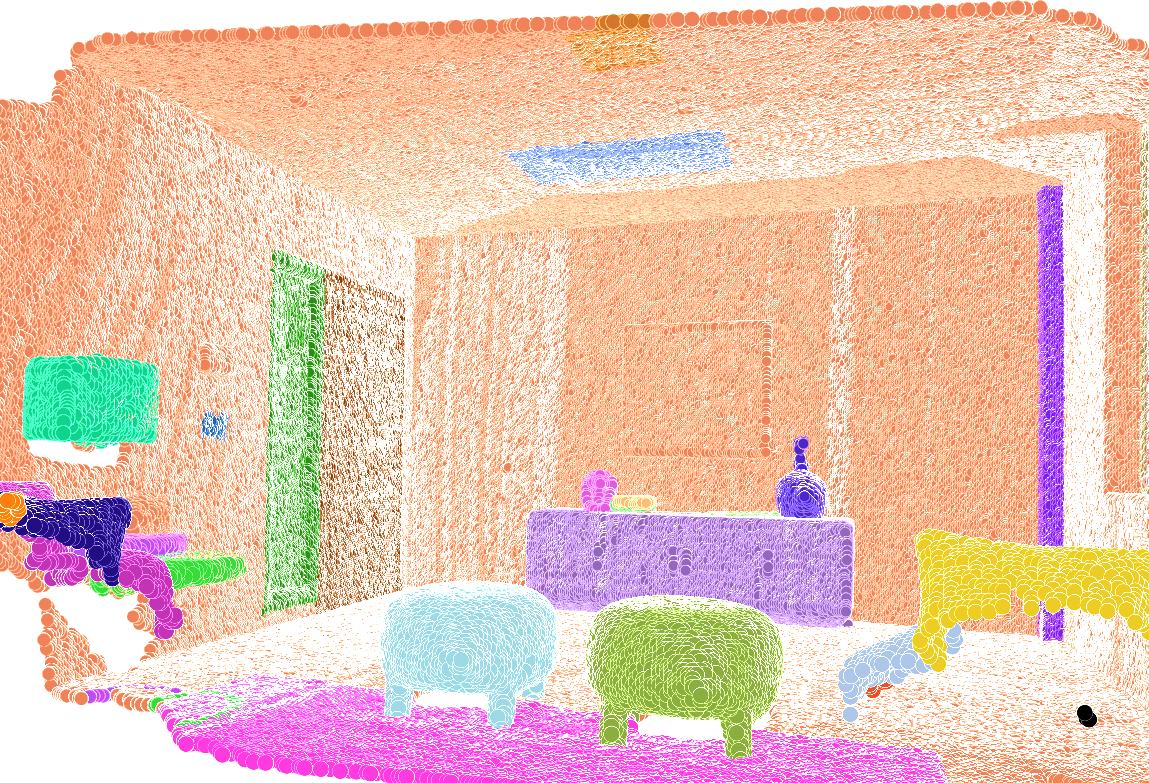}
                \put(40,-7){\color{black}\textbf{Ground-truth}}
            \end{overpic}
        &
        \begin{overpic}[width=0.235\textwidth]{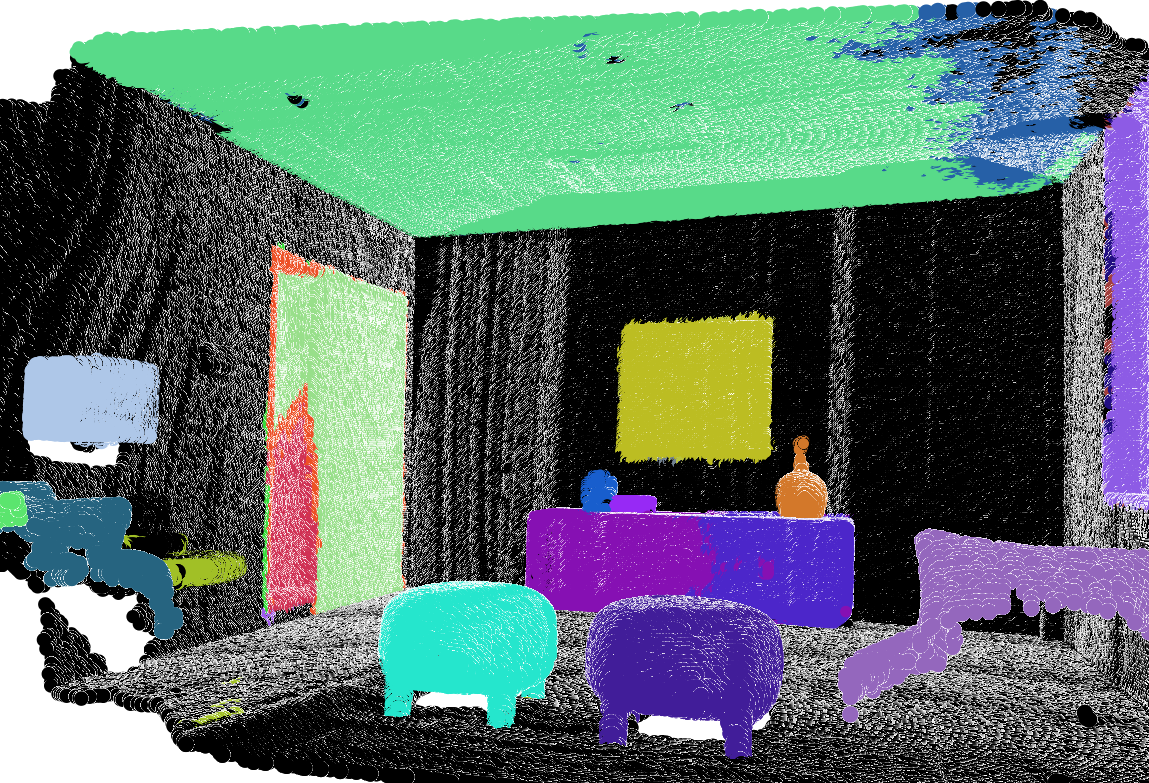}
            \put(40,-7){\color{black}\textbf{Prediction}}
        \end{overpic}
            \\
    \end{tabular}
    \vspace{1mm}
    \caption{Qualitative results were obtained using \ourmethod in the \taskname setting on Replica. The process begins by generating 2D instance masks for each image by querying a vocabulary established by LLaVA (as shown in the top row, where the words are displayed in each image). These 2D masks (highlighted by boxed regions in each image) are then integrated into 3D masks (bottom row: Prediction columns) using spectral clustering techniques.
    }
    \label{fig:supp_replica}
\end{figure*}

Fig.~\ref{fig:supp_replica} presents qualitative results from two examples of instance segmentation achieved using our method in the \taskname setting. For each example, we provide 2D instance segmentation results from two different view RGB images of a 3D scene, 3D instance ground truth and our prediction. Our method demonstrates visually robust performance without relying on predefined categories.

\subsection{Visualizations on ScanNet200.}
\begin{figure*}[t]
\centering
\tabcolsep 1pt
    \begin{tabular}{cccc}
        \begin{overpic}[width=0.235\textwidth]{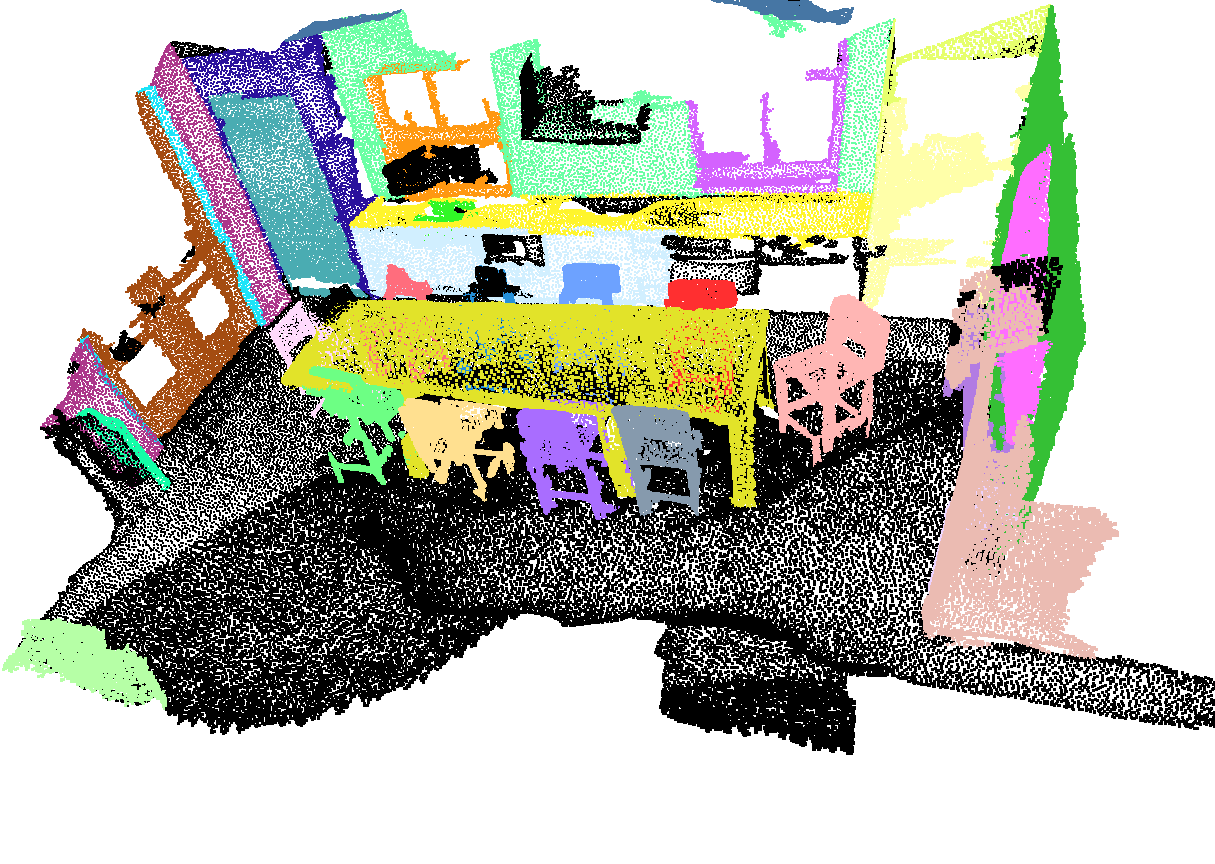}
            \put(35,72){{\footnotesize \textbf{Ins. GT}}}
            \end{overpic}      
        & 
        \begin{overpic}[width=0.235\textwidth]{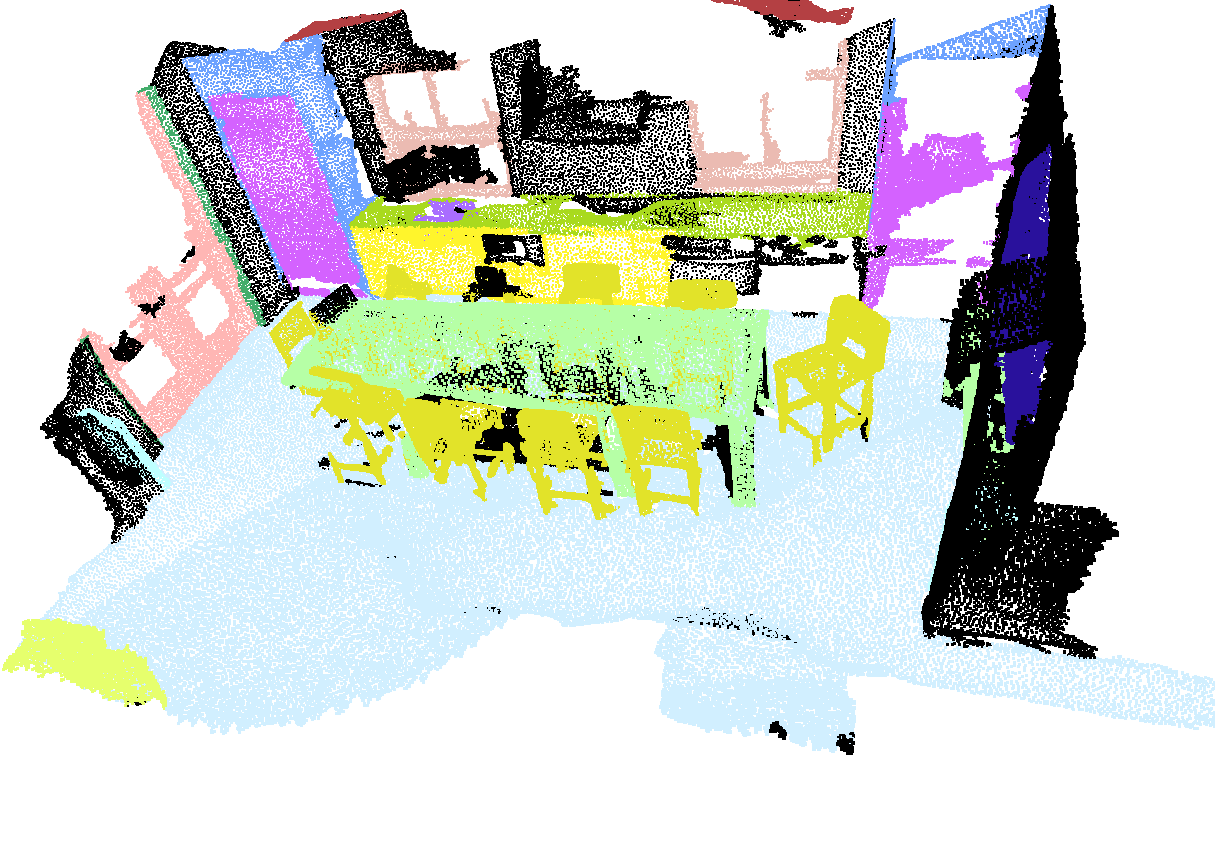}
            \put(35,72){{\footnotesize \textbf{Sem. GT}}}
            \end{overpic}
        &
        \begin{overpic}[width=0.235\textwidth]{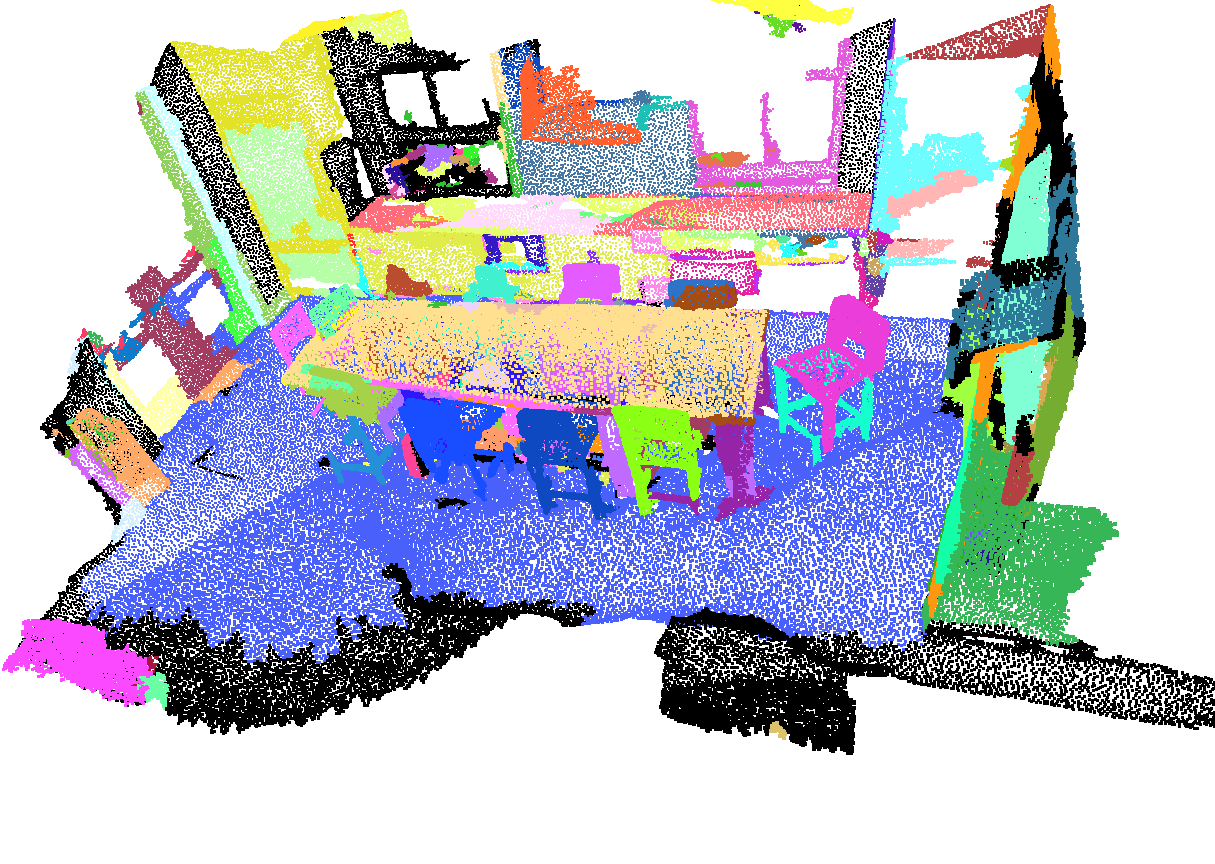}
            \put(35,72){{\footnotesize \textbf{Ins. Pred.}}}
            \end{overpic}
        &
        \begin{overpic}[width=0.235\textwidth]{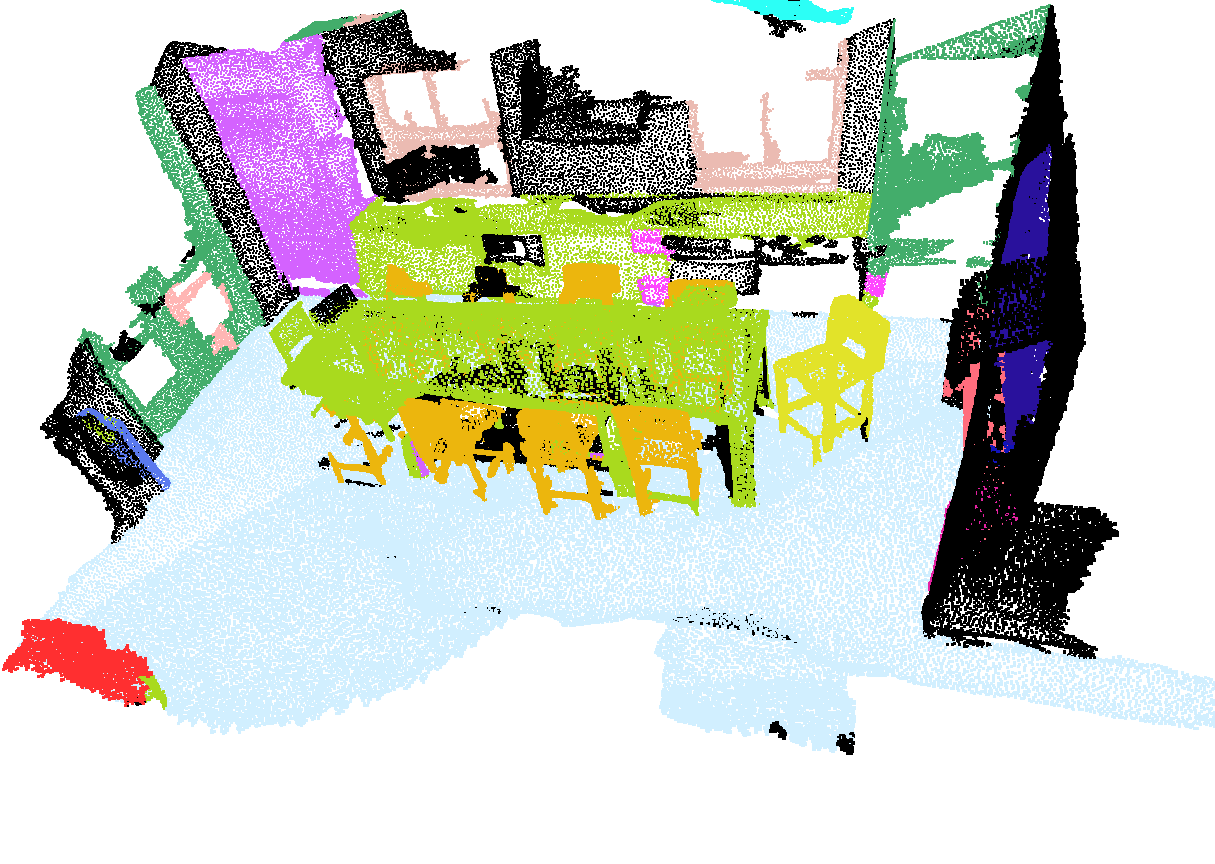}
            \put(35,72){{\footnotesize \textbf{Sem. Pred.}}}
            \end{overpic}
        \\
        \begin{overpic}[width=0.235\textwidth]{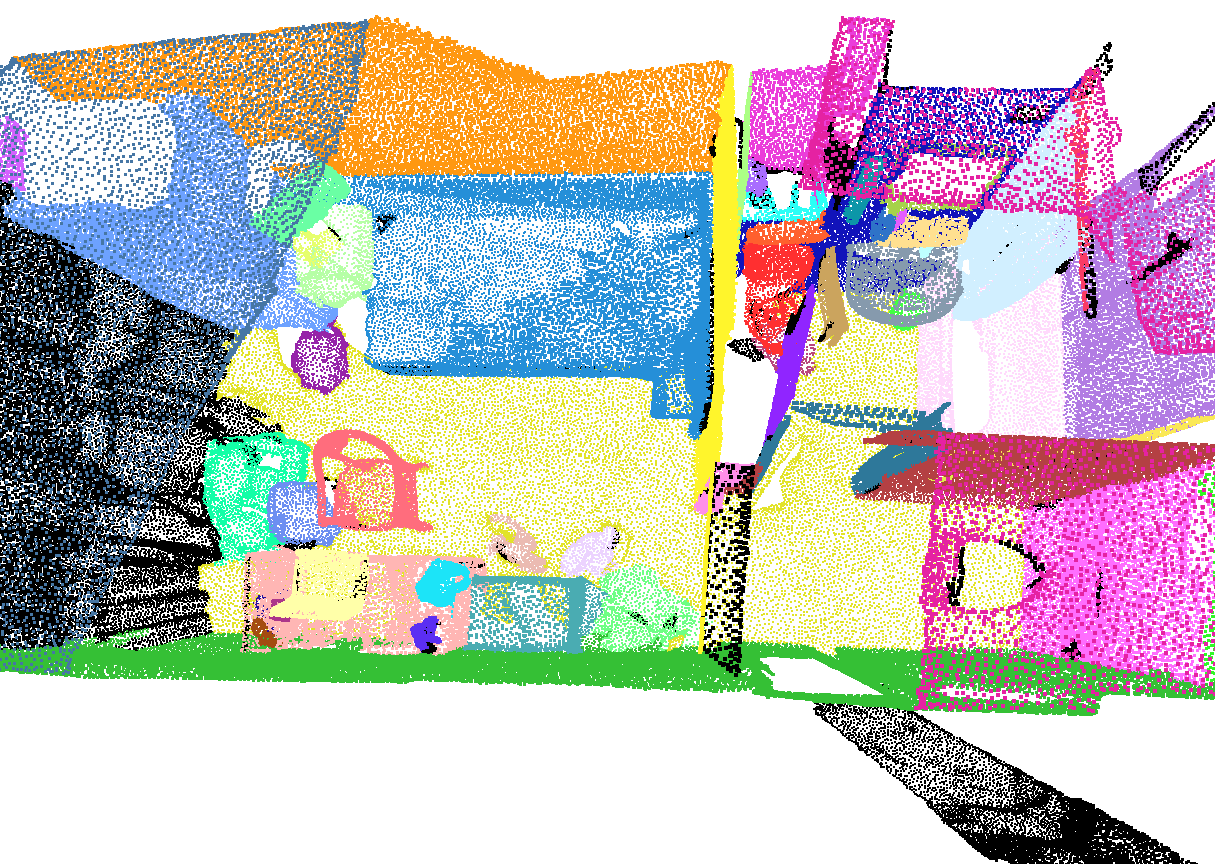}
            \end{overpic}      
        & 
        \begin{overpic}[width=0.235\textwidth]{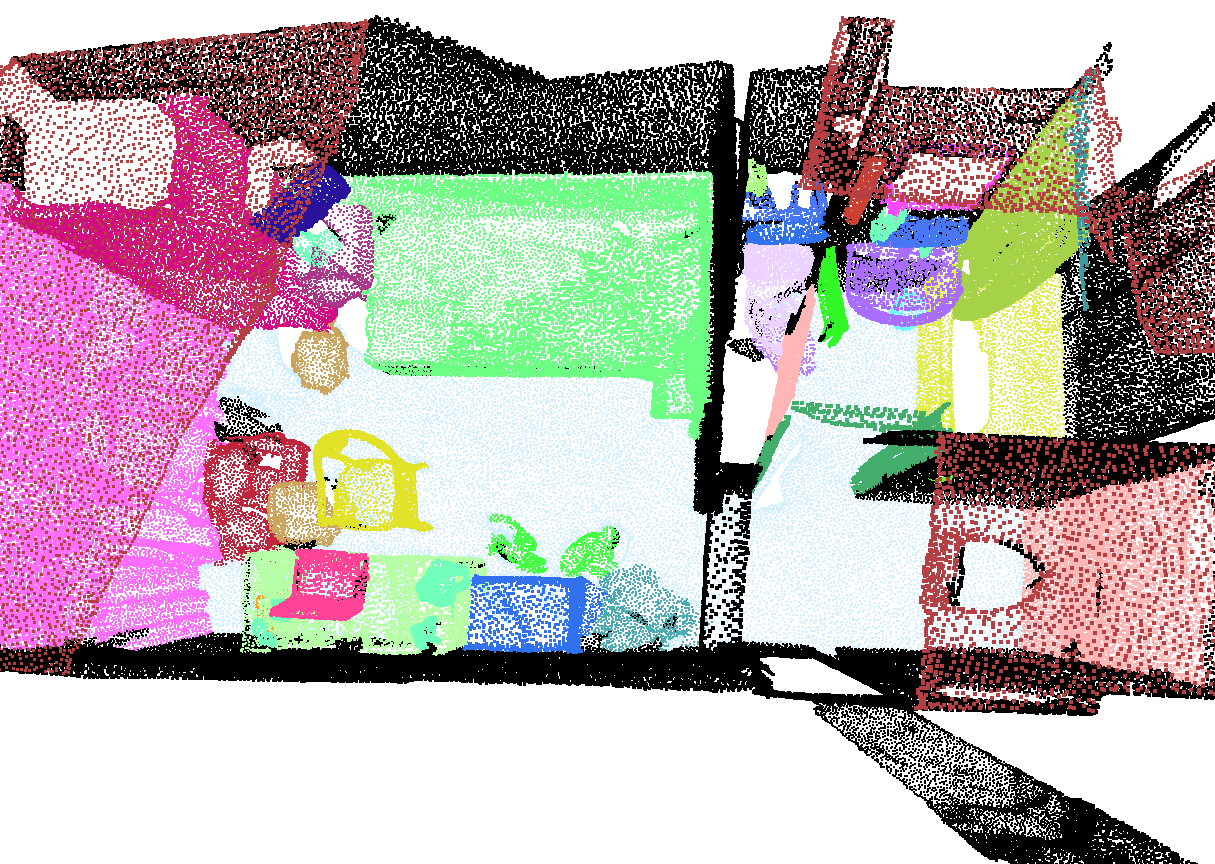}
            \end{overpic}
        &
        \begin{overpic}[width=0.235\textwidth]{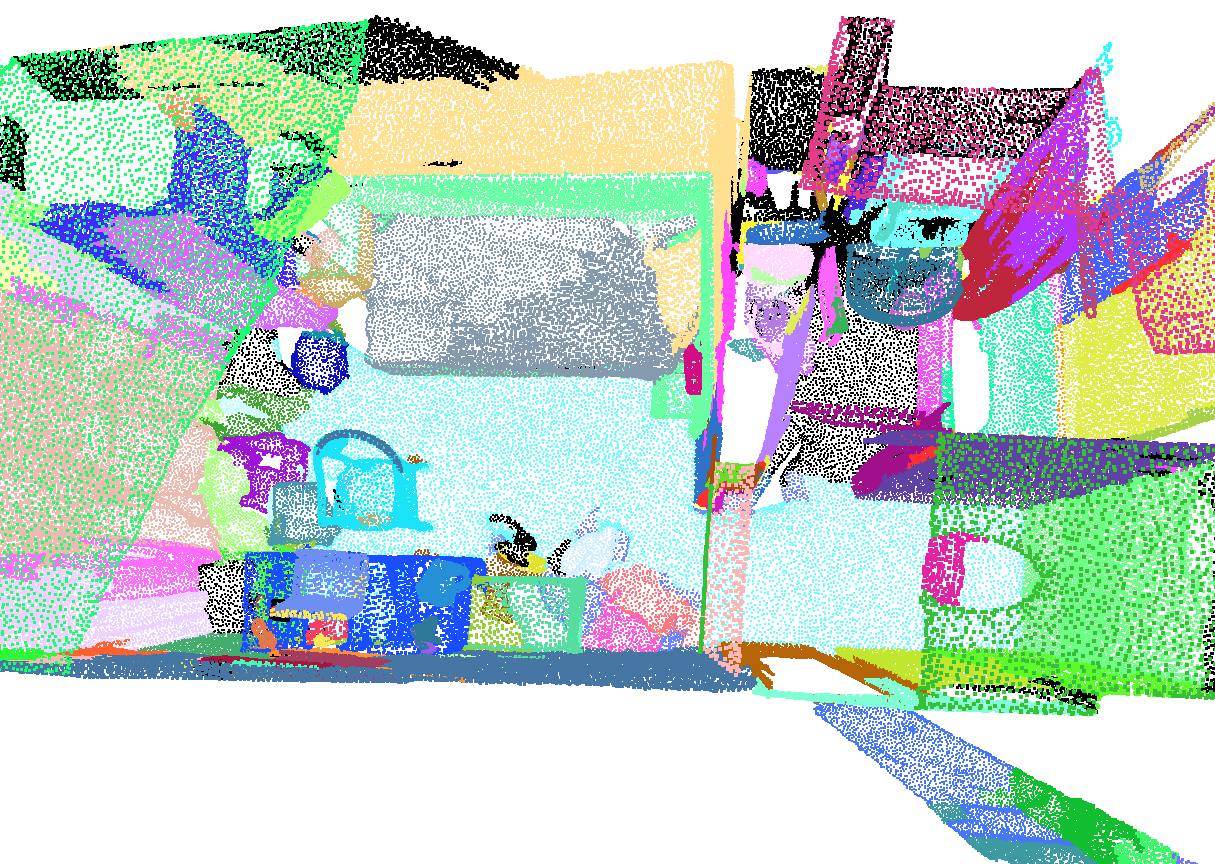}
            \end{overpic}
        &
        \begin{overpic}[width=0.235\textwidth]{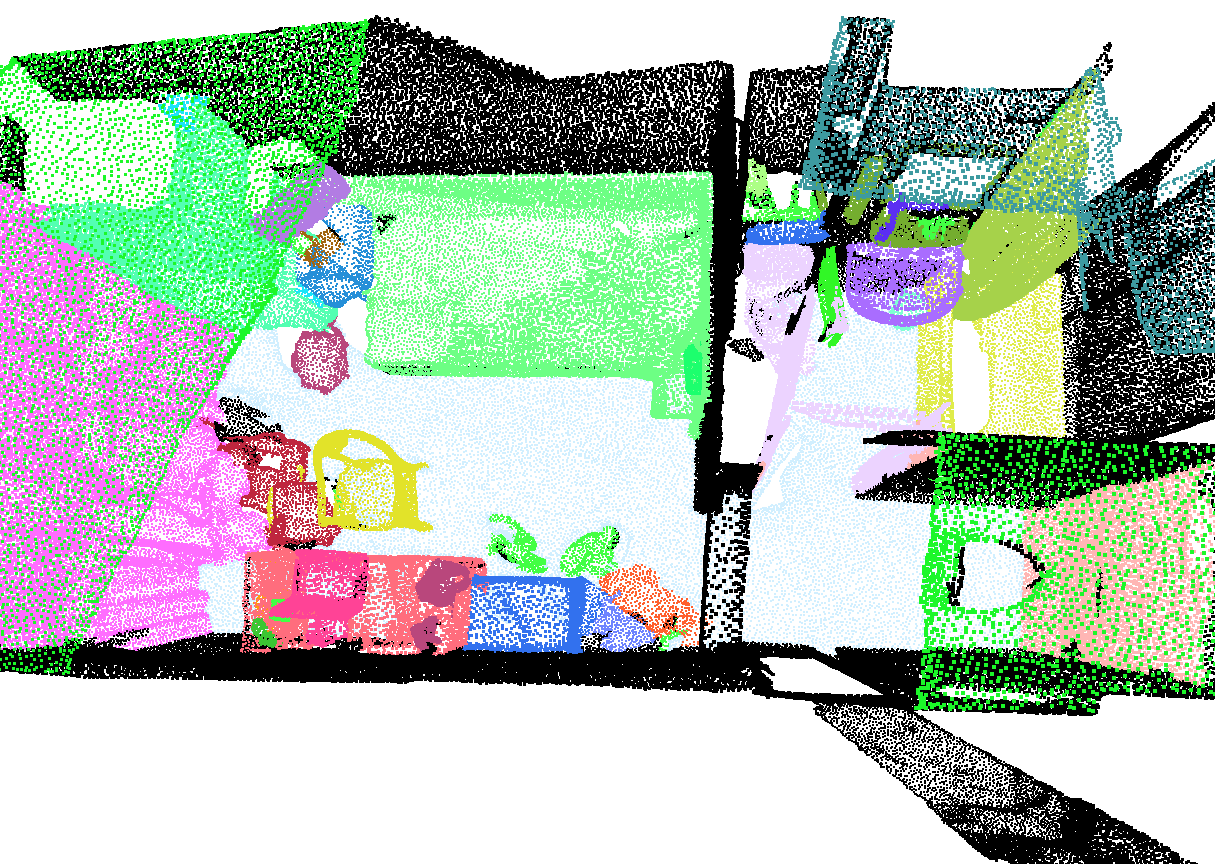}
            \end{overpic}
        \\
        \begin{overpic}[width=0.235\textwidth]{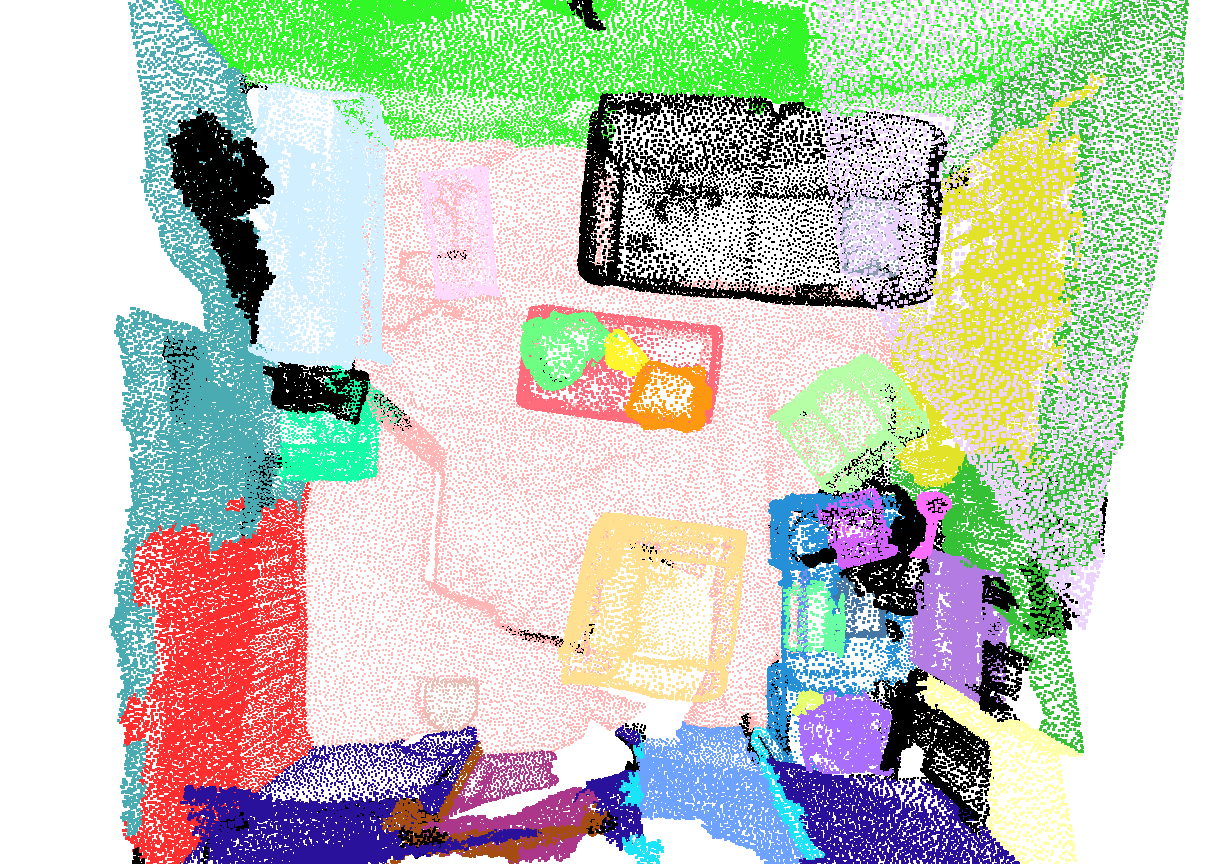}
            \end{overpic}      
        & 
        \begin{overpic}[width=0.235\textwidth]{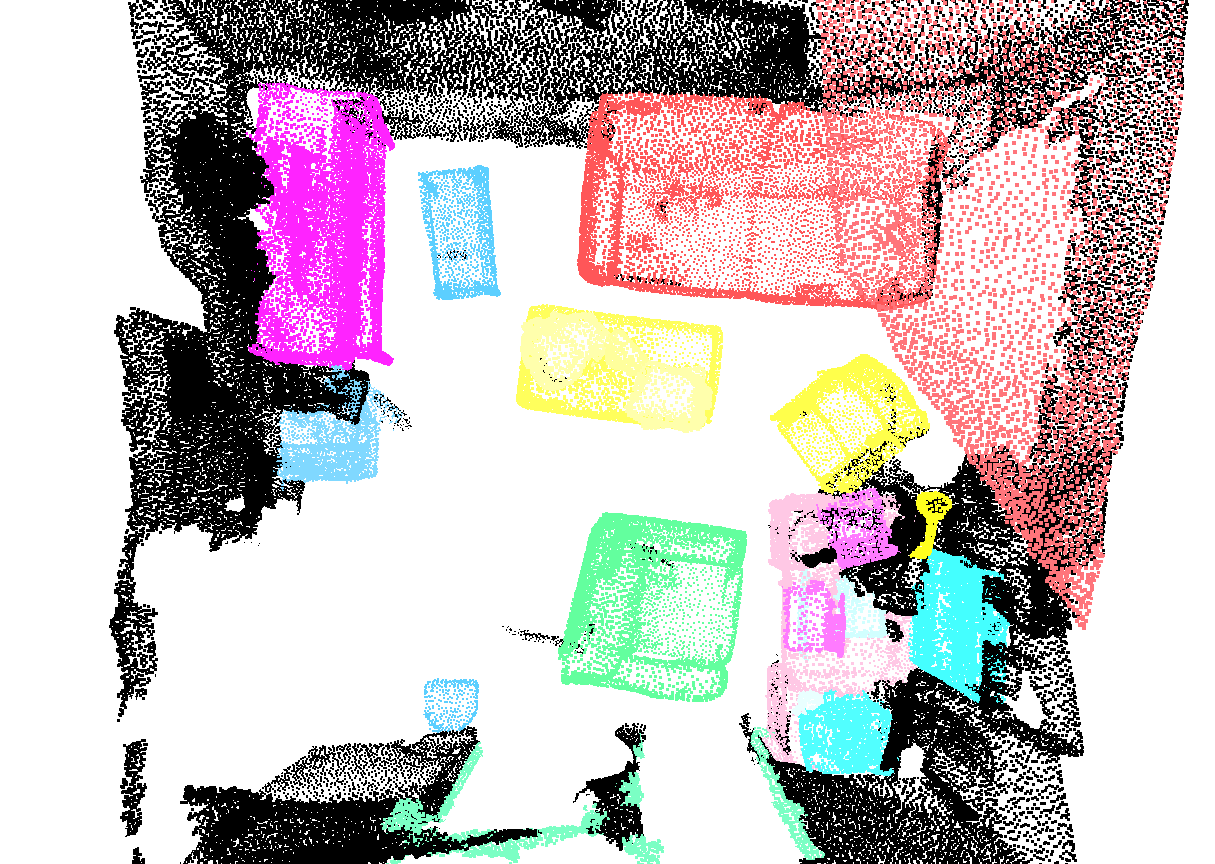}
            \end{overpic}
        &
        \begin{overpic}[width=0.235\textwidth]{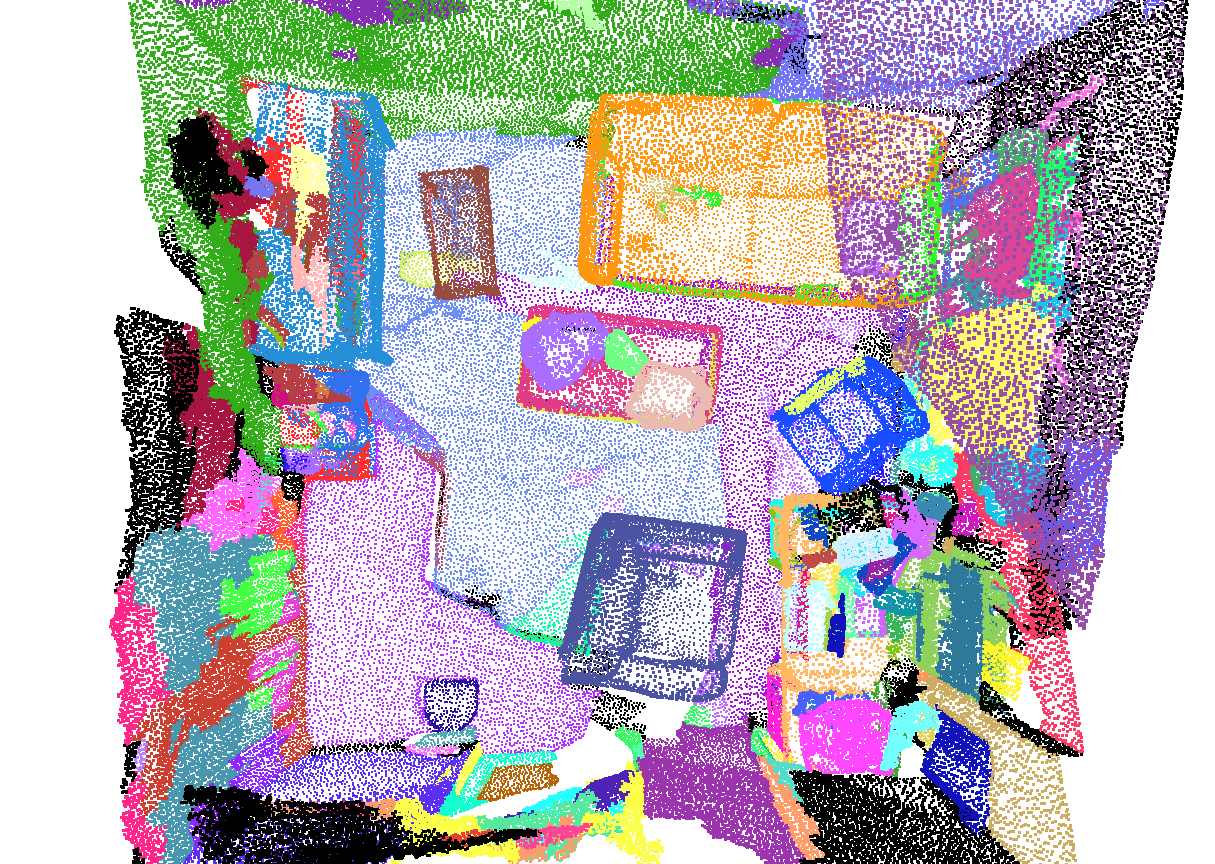}
            \end{overpic}
        &
        \begin{overpic}[width=0.235\textwidth]{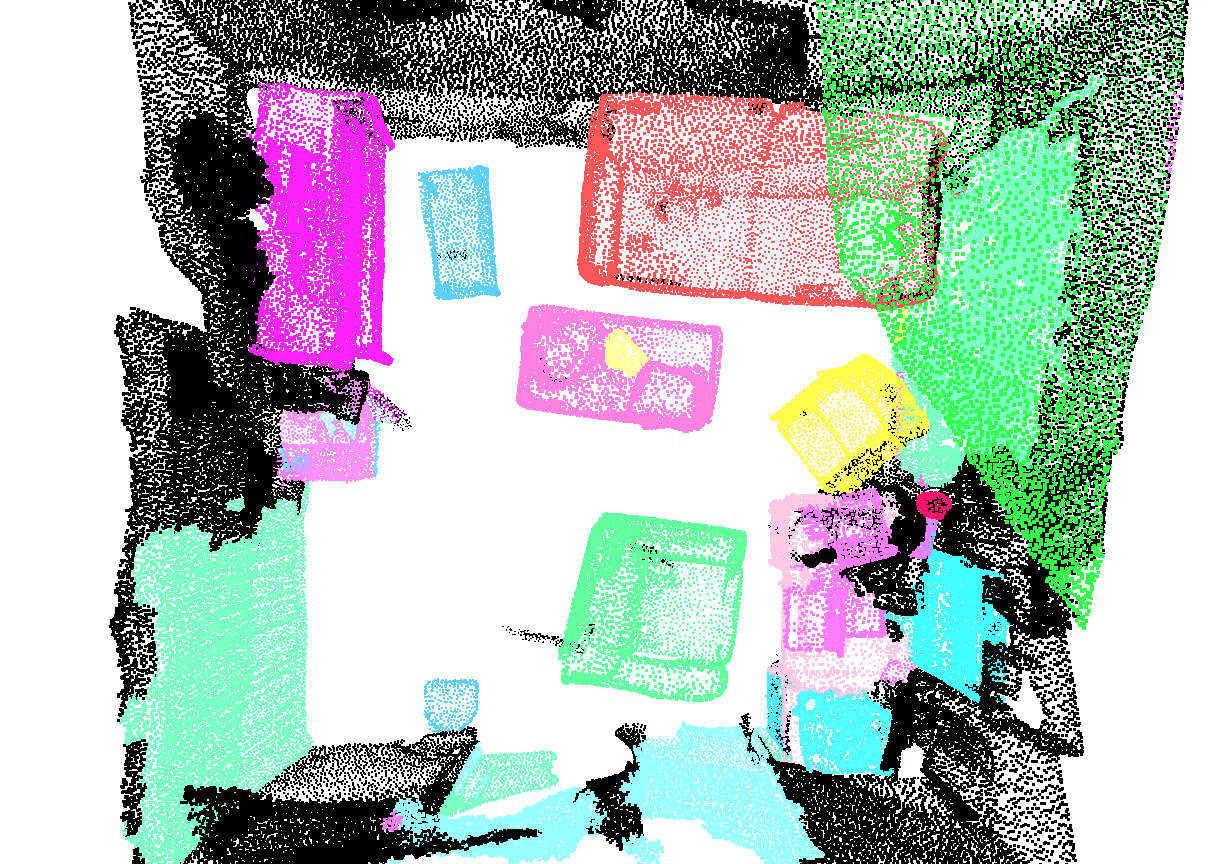}
            \end{overpic}
        \\
        \begin{overpic}[width=0.235\textwidth]{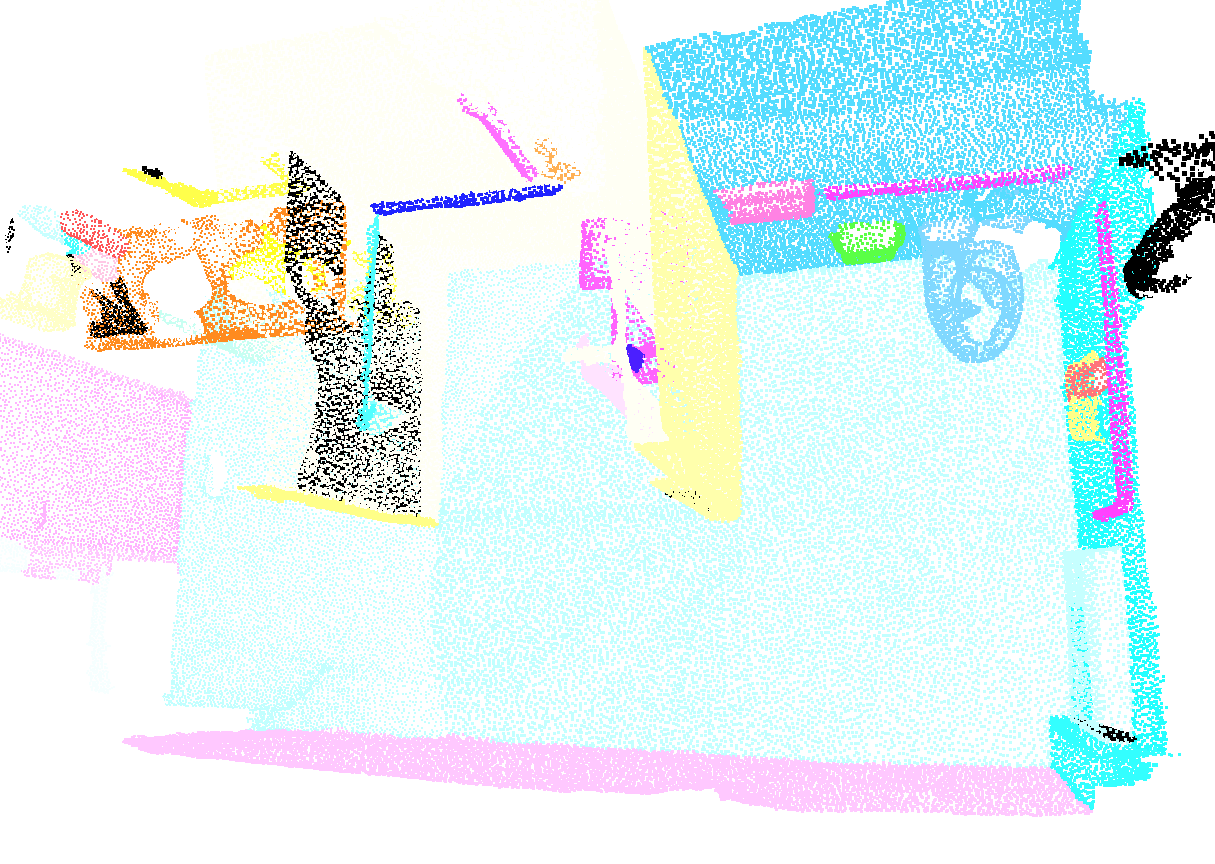}
            \end{overpic}      
        & 
        \begin{overpic}[width=0.235\textwidth]{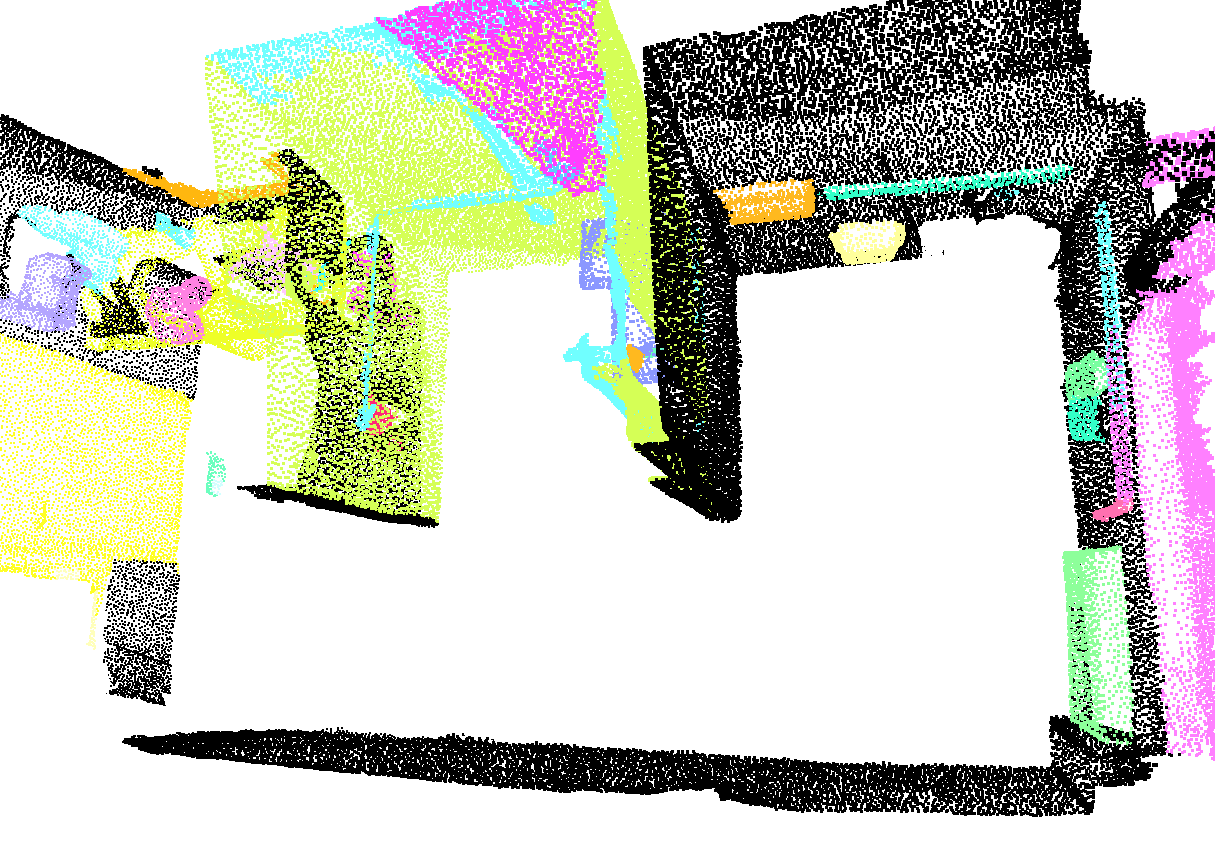}
            \end{overpic}
        &
        \begin{overpic}[width=0.235\textwidth]{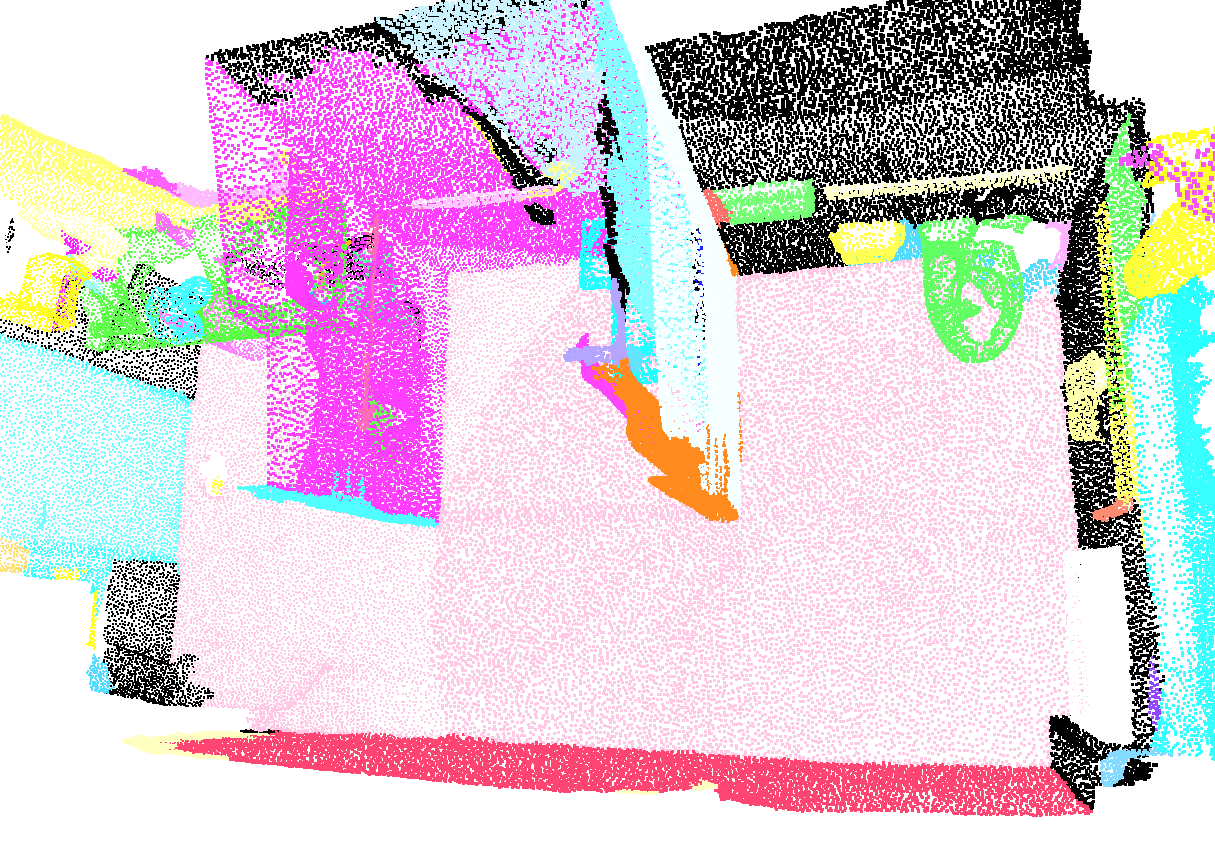}
            \end{overpic}
        &
        \begin{overpic}[width=0.235\textwidth]{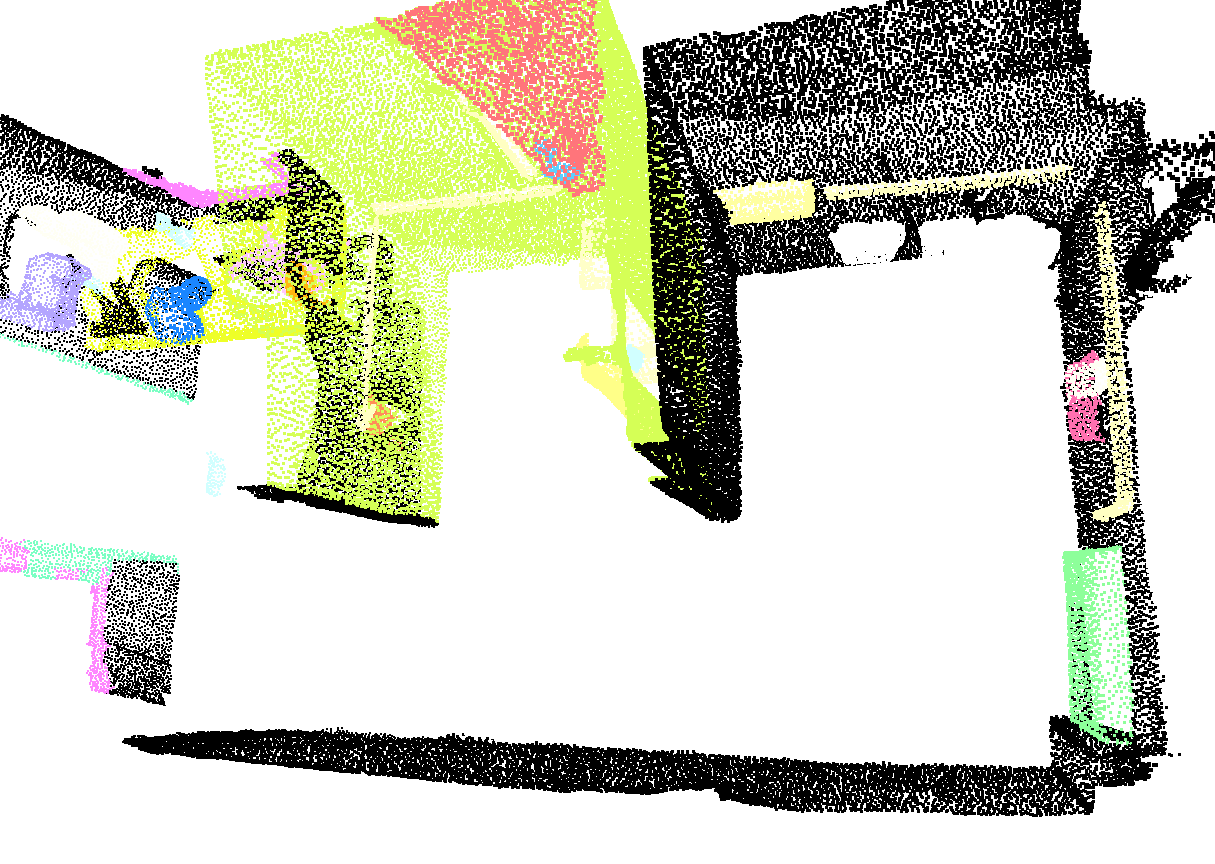}
            \end{overpic}
        \\
    \end{tabular}
    \vspace{-3mm}
    \caption{Qualitative results obtained by \ourmethod in the \taskname setting on ScanNet200 are presented. From left to right: ground truth instance labels, ground truth semantic labels, predicted 3D instance labels, and predicted 3D semantic labels. In ideal instance segmentation within a 3D scene, different instances should be colored differently, while the same instance should have a consistent color. It is not necessary for the same object to have the same color in both the ground truth and predicted results. For semantic prediction, success is indicated when each object matches the ground truth color.
    }
    
    \label{fig:supp_scan}
\end{figure*}

To illustrate the quality of segmentation, we provide additional visualizations on ScanNet200.
Fig.~\ref{fig:supp_scan} presents four examples of instance segmentation results.  
Ideally, different instances should have distinct colors, while the same instance should maintain consistent coloring. It's not necessary for objects to match colors exactly between the ground truth and predictions, but semantic success is achieved when objects match the ground truth colors. As shown, \ourmethod accurately segments most of the scene for both instance and semantic segmentation using only the LLaVA-provided vocabulary.

\section{Implementation details for \taskname setting}
In our experiments, because the categories provided by the annotated datasets are fewer than the actual objects they contain, we evaluate our method in the \taskname setting by restricting it to the categories detected by both LLaVA and those provided in the datasets.

\end{document}